\documentclass{article}

\usepackage[nonatbib, preprint]{neurips_2023}

\usepackage[pagebackref=true,breaklinks=true,letterpaper=true,colorlinks,bookmarks=false]{hyperref}

\usepackage[utf8]{inputenc} 
\usepackage[T1]{fontenc}    
\usepackage{hyperref}       
\usepackage{url}            
\usepackage{booktabs}       
\usepackage{amsfonts}       
\usepackage{nicefrac}       
\usepackage{microtype}      
\usepackage{xcolor}         
\usepackage{graphicx}
\usepackage{epsfig}
\usepackage{graphicx}
\usepackage{amsmath}
\usepackage{amssymb}
\usepackage{comment}
\usepackage{tabularx}
\usepackage{tikz}

\DeclareMathOperator{\Tr}{Tr}
\DeclareMathOperator*{\argmin}{arg\,min}
\usepackage{subcaption}

\title{Registering Neural Radiance Fields as \\ 3D Density Images}

\author{%
  Han Jiang\thanks{Equal contribution. The order of authorship was determined alphabetically.} \\
  HKUST\\
  \texttt{hjiangav@connect.ust.hk} \And
  Ruoxuan Li\footnotemark[1] \\
  HKUST\\
  \texttt{rliba@connect.ust.hk} \And
  Haosen Sun \\
  HKUST\\
  \texttt{hsunas@connect.ust.hk} \And
  Yu-Wing Tai \\
  HKUST\\
  \texttt{yuwing@gmail.com} \And
  Chi-Keung Tang \\
  HKUST\\
  \texttt{cktang@cs.ust.hk}
}

\begin{document}

\maketitle

\begin{abstract}
No significant work has been done to directly merge two partially overlapping scenes using NeRF representations.
Given pre-trained NeRF models of a 3D scene with partial overlapping, this paper aligns them with a rigid transform, by generalizing the traditional registration pipeline, that is, key point detection and point set registration, to operate on 3D density fields. 
To describe corner points as key points in 3D, we propose to use universal pre-trained descriptor-generating neural networks that can be trained and tested on different scenes. We perform experiments to demonstrate that the descriptor networks can be conveniently trained using a contrastive learning strategy. We demonstrate that our method, as a global approach, can effectively register NeRF models, thus making possible future large-scale NeRF construction by registering its smaller and overlapping NeRFs captured individually.
\end{abstract}

\section{Introduction}
\label{sec:1}

3D scene registration is a fundamental task for constructing large-scale 3D scenes, with numerous important applications such as virtual reality and panoramic indoor and outdoor maps. Research efforts have been made on registering traditional explicit 3D scene representations including point clouds ~\cite{park2017colored, sift_point, 3dfeat} and meshes ~\cite{multires_mesh, mesh_dim}, which have achieved good results. On the other hand, Neural Radiance Fields (NeRF) ~\cite{NeRF_2020} provide a novel implicit scene representation that generates images of 3D scenes by volume rendering. The rapid development of NeRF in recent years has revealed its high potential and has made it the next typical 3D scene representation. This demands a method for NeRF registration that this paper focuses on.

To register NeRFs, we may directly operate on the continuous neural field, or convert NeRF into existing discrete scene representations. While directly dealing with continuous fields is natural and expected to be accurate, the implicit nature of neural continuous fields introduces complexity to the problem, where frequent and irregular queries to NeRF are expected. Conversion to explicit scene representations is easier and more convenient during registration, but the conversion itself is not obvious and may introduce additional inaccuracy to registration. On the other hand, the problem of traditional 2D image registration has been extensively studied, for which there exists a mature pipeline: key point detection, key point description, descriptor matching and registration. In essence, a 2D image is a discrete representation of RGB color fields, and the goal of image registration is to align two color fields. Inspired by 2D image registration, in this paper, we propose to convert NeRF into 3D images rather than other scene representations, and perform registration on 3D images. In order to make our registration photometric-invariant, we target at registering the geometry of the scene only, which correspond to density fields in NeRF. Thus we only make use of neural density fields, but discard radiance fields to avoid disturbance from illumination. Hence our registration framework only needs to query NeRF density, and we only need a one-time query on 3D image texels, i.e., grid nodes. Therefore, the conversion to 3D density images is efficient and takes advantage of explicitness. In addition, 3D images can be easily downsampled into multi-scale, which makes our registration framework scale-invariant.

Similar to the image registration pipeline, good key point descriptors are critical in our framework. Designing 3D descriptors are more challenging than 2D due to the increasing complexity of corner appearances in 3D. In view of the success of neural descriptors on 2D image features~\cite{cnn-sift} and other 3D representations~\cite{neural_descriptor_fields,ExMeshCNN}, we choose to use a universal neural network that generates rotation-invariant descriptors from 3D density image patches extracted from any scene. The network is expected to generate descriptors good enough for matching and registration without the need for fine-tuning, thus allows efficient descriptor generation. Despite the universality and convenience of this network, its training does not require much data effort. Specifically, we detect corners in various training scenes and sample their local neighborhoods in several different orientations, which synthesizes a large amount of training data. Then we propose a contrastive learning strategy that effectively trains our universal network.

Our contributions mainly consist of two parts: 1) we propose, to the extent of our knowledge, the first 3D density image based NeRF registration framework; 2) we propose a universal neural 3D corner descriptor, coupled with a strategy to train this network with contrastive learning.

\section{Related Work}
\label{sec:2}

\subsection{NeRF}
Neural radiance field (NeRF)~\cite{NeRF_2020} is a revolutionary method with high potential for novel view generation and 3D reconstruction, which utilizes an implicit radiance field to model a certain given scene. More works are done on this original NeRF representation idea. Instant neural graphics primitives (instant-ngp)~\cite{mueller2022instant} provides state-of-the-art optimization in the training process. The training time for one single NeRF drops dramatically from around 10 hours to less than 10 seconds by applying multi-resolution hash encoding. There are also other works depending on NeRF exploring different perspectives. NeRF-RPN~\cite{NERF-RPN}, makes a great contribution to object detection in the radiance field. Nerf2nerf~\cite{nerf2nerf} focuses on high-quality object registration in different NeRF scenes. Works are also done for optimizing the training process on a large scale by Mega-NeRF~\cite{Mega-nerf} and Block-NeRF~\cite{Block-nerf}. Although the mentioned works have touched on the topic of NeRF registration, none of them has contributed in registering two overlapping NeRF scenes.

\subsection{NeRF and 3D Registration}

To make future NeRF research applicable to complex and large-scale starting with indoor scenes, registering overlapping NeRFs captured at different times and possibly resolution is a necessary step. To date, however, there is sparse research effort to address this fundamental 3D computer vision problem for NeRFs. Bundle-adjusting neural radiance fields (BARF) and its variants ~\cite{lin2021barf, chen2022local, kim2022visual} contribute to the registration of camera poses by learning the 3D scene representation of the original NeRF. Zero NeRF~\cite{peat2022zero} leverages the NeRF representation to register two sets of images with almost no correspondence. Nerf2nerf~\cite{nerf2nerf} studied the pairwise object registration in different NeRF scenes, which helps in semantic analysis in the NeRF space. However, none of the mentioned works aimed at large-scale 3D NeRF registration. That is, there is still no solution to merging two overlapping scenes directly using NeRF representation.
On the other hand, 3D scene registration has been extensively researched in computer vision, mostly relying on explicit and discrete representation, unlike NeRFs. In particular, point cloud registration has been extensively studied. Most point cloud registration methods depend heavily on explicit features to localize the points. Some commonly used features are the shapes~\cite{wohlkinger2011ensemble} and point feature histogram (PFH)~\cite{rusu2009fast}. Another notable 3D approach consists of registering meshes. Typically, due to the expensive 3D computation, 3D meshes are either downsampled ~\cite{multires_mesh} or reduced in dimension ~\cite{mesh_dim} to avoid the expensive computation.  We are inspired by the ideas of rigid registration of 3D point clouds to avoid heavy computation. But these methods are limited to their input, which differs from the continuous implicit 3D representation of NeRF. Thus, none of them can be directly utilized for 3D NeRF registration.

\subsection{Feature Descriptors}
Hand-crafted non-learning-based descriptors, represented by SIFT and its variants~\cite{2dsift, chiu2013fast, abdel2006csift, liu2008sift, ke2004pca, morel2009asift, nsift} and an array of well-known descriptors (e.g., HOG~\cite{dalal2005histograms}, SURF~\cite{bay2006surf}, MOPS~\cite{brown2005multi}, ORB~\cite{rublee2011orb} etc), have been extended to 3D SIFT~\cite{3dsift} and employed in matching 2D and 3D imageries in a wide range of applications, such as 2D/3D medical registration~\cite{kumar2009predicting, allaire2008full}, nonrigid mesh registration~\cite{sift_mesh}, point cloud registration~\cite{sift_point}, RGB-D registration~\cite{sift_rgb} to name a few. SIFT is still widely adopted due to its high robustness and invariance.
Learning-based descriptors, such as~\cite{self_corner_detection, 3dmatch, s_sift, 3dfeat, ppf} have been proposed. While excellent results have been reported in  matching the above {\em discrete} domains, none of them are designed to match 3D continuous density volumes which are  different from discrete point cloud (clustered), mesh (irregular), and RGB-D (only quasi-dense) data. 
This paper regards NeRFs as {\em 3D images}, and adopts a data-driven approach to learning 3D neural descriptors at detected corners in NeRF density fields. Similar to SIFT~\cite{2dsift,3dsift}, by construction, our neural descriptors are photometric, scale, and rotational invariant, which will be detailed in the next section. 

With the development of machine learning, network-based descriptors started to outperform traditional hand-crafted ones. Unsupervised convolutional neural networks perform far better than the classic SIFT algorithm in descriptor-matching tasks~\cite{cnn-sift}. Based on the capability of CNN, many improved network architectures were proposed~\cite{zhang2017learning, hard-net, kernel-sp, local-ap, matchnet}. These deep-learning-based descriptors are extensions of the hand-crafted descriptors in some way~\cite{des-compare} because most of them depend on classical algorithms. Due to the power of deep neural networks, the performance is highly improved. We further implement a 3D contrastive learning descriptor in the NeRF field in order to be compatible with the implicit spatial density representation.

\section{Method}
\label{sec:3}

Given two neural radiance and density fields $\mathit{F}_1: (\mathbf{x}_1 , \mathbf{d})\rightarrow (\mathbf{c}, \sigma)$ and $\mathit{F}_2: (\mathbf{x}_2 , \mathbf{d})\rightarrow (\mathbf{c}, \sigma)$, where $\mathbf{x}_1\in V_1 , \mathbf{x}_2\in V_2$ such that $V_1, V_2$ are two overlapping volumes  (i.e., $V_1 \cap V_2 \neq \emptyset$), we aim to solve for an optimal rigid transformation making the two fields align with each other. This rigid transformation can be represented as $\mathbf{x}_1 = lR\mathbf{x}_2 + \mathbf{t}$, where $l \in \mathbb{R}^{+}$ is a scale factor, $R$ is a $3 \times 3$ rotation matrix with 3 degrees of freedom, and $\mathbf{t}$ is a 3D translation vector.

With density grids, i.e. 3D density images, extracted from the two NeRFs to be registered, our registration pipeline generalizes 2D image registration to 3D, where we perform matching using neural 
feature descriptors. First, we discretize the two NeRFs to density grids with filtering strategies that eliminate noise from sampling, and then downsample them to get multi-scale density grids. Second, we operate a 3D version of Harris corner detector on the grids to compute Harris response in multi-scale, followed by non-maximum suppression to determine corner point locations. Next, we extract density grids from multi-scale neighborhoods of the corners. These neighborhood density grids are fed into a pre-trained neural descriptor network to generate corner descriptors, which are then matched to obtain correspondences of those corners. Finally, we use RANSAC to compute the rigid transformation between the two corner point sets, and regard this transformation as our solution.

We will frequently use notations of grids to represent related computations. In this section, all arithmetic operations on grids are interpreted as element-wise operations on each grid cell value. Also, although we describe our registrtion as rigid, our registration is not strictly rigid because we additionally consider a scale factor $l$. This is considered since NeRFs to be registered can be in different scales.

\subsection{Neural Density Field Discretization}
\label{sec:3.1}
Our registration pipeline requires corners as key points for matching. To detect corners, analogous to 2D Harris corner detectors on images, we seek to adapt 3D Harris detectors on NeRFs. NeRF, more than its name suggests, represents a radiance field as well as a density field. While the radiance field depends on viewing directions and does not separate color and illumination, the density field represents scene geometry in NeRF and is only related to query positions. Therefore, to robustly find corners of scene geometry without being affected by environmental lighting, we only need to extract and pre-process density fields.

We discretize the continuous neural field using a grid covering the whole scene. Extracting grids from the continuous 3D field is essentially sampling signals in 3D spaces. In the meantime, NeRFs obtained from indoor scenes are often too noisy for the purpose of corner detection, due to the relatively insufficient training images. Thus, we need an appropriate sampling method to make our samples represent the continuous density signals well. 

Directly sample densities on each grid node locations is expected to extract a lot of noise from the continuous field. To filter out noise, we may first sample with a high resolution grid, and then downsample with smoothing operations, such as average pooling and Gaussian filter. However, these types of filters is known to smooth edges and corners, making our downstream corner detection task more difficult. Therefore, we choose to denoise our sampled density grids with techniques that preserve corners as large variations in densities, such as anisotropic diffusion ~\cite{anisotropic}. We provide further discussion in section~\ref{sec:4.4} about the advantages of this type of techniques.

\begin{figure}[t]
\begin{center}
   \includegraphics[width=1\linewidth]{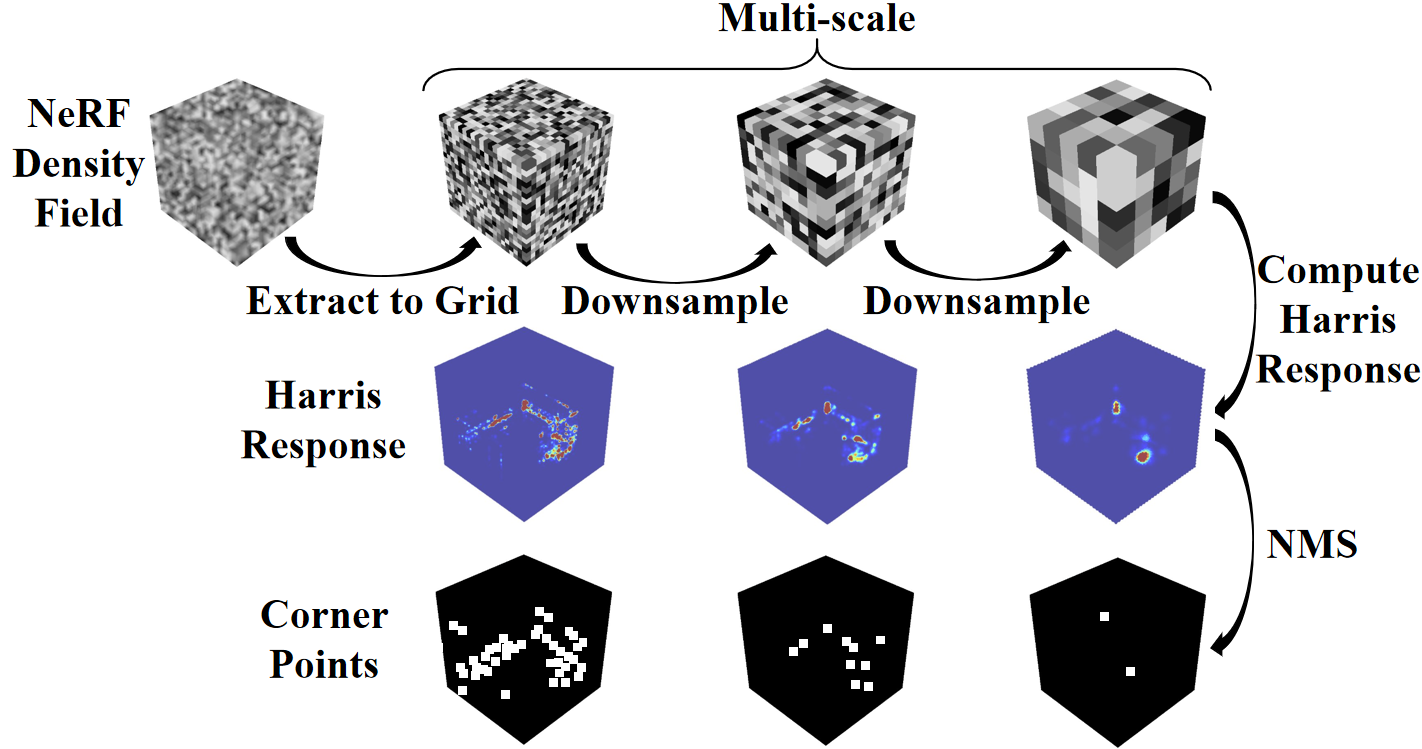}
\end{center}
   \caption{Multi-scale Corner Detection on a 3D Density Image.}
\label{fig:1}
\end{figure}

\subsection{Corner Detection}
\label{sec:corner_detect}

We first extract high-resolution density grids $G^0_1, G^0_2$ from the NeRF pair, and regard each of them as a 3D image. As Harris detectors are not scale-invariant, as shown in in Figure~\ref{fig:1}, we downsample $G^0_1, G^0_2$ with blurring filters to filter out high-frequency geometry information. This generates two sets of multi-scale grids $\{G^0_1, G^1_1, \dots, G^d_1\}, \{G^0_2, G^1_2, \dots, G^d_2\}$, where the number of scales $d$ is manually determined according to the scene. We typically use $d = 3$ or $4$. Discretizing the original continuous neural density field into multi-scale grids not only facilitates density queries, which makes later steps of registration convenient, but also takes large-scale, low-frequency geometry information into consideration.

For each grid $G$, we then use a 3D operator to compute density gradients $I_x, I_y, I_z$ in three directions. For example, a 3D Sobel operator, defined as $(1, 0, -1) \otimes (1, 2, 1) \otimes (1, 2, 1)$ where the symbol $\otimes$ denotes outer product, can be applied here.  Then we form a Harris matrix for each grid cell, which constructs a Harris matrix grid
\begin{equation}
\label{eq:harris_matrix}
M = \sum_{W}
\begin{pmatrix}
I_x I_x & I_x I_y & I_x I_z\\
I_x I_y & I_y I_y & I_y I_z\\
I_x I_z & I_y I_z & I_z I_z
\end{pmatrix}
\end{equation}
where $W$ denotes local 3D windows around grid cells. With $M$, 3D Harris response grid $H$ can be computed by
\begin{equation}
\label{eq:harris_response}
H = \det M - k (\Tr M)^2
\end{equation}
where $k$ is a manually chosen hyper-parameter. We typically use $k=0.06$.

Once the global response grids $H$ is obtained for each scene and scale, we  perform non-maximum suppression (NMS) on $H$ to find its local maxima as corner point locations in the two scenes. Before NMS, we may filter out low response values because local maxima with very low response are highly likely to be noises. Grid indices of corners detected in every scale are then converted into coordinates to construct two sets $X_1, X_2$ for each scene for future use. This step is visualized as Figure~\ref{fig:1}.

\subsection{Rotation-invariant Neural Descriptor}

The next step in our pipeline is to create descriptors for all corners. As the two NeRFs to be registered may be trained under different scales, poses and lighting, our corner descriptors are supposed to be invariant to these factors to support accurate matching. Since our corner detection is of multi-scale and only operates on density, our descriptors are expected to be scale and illumination invariant already. Thus we focus on developing a strategy for rotation-invariant descriptors.

\begin{figure}[t]
\begin{center}
   \includegraphics[width=1\linewidth]{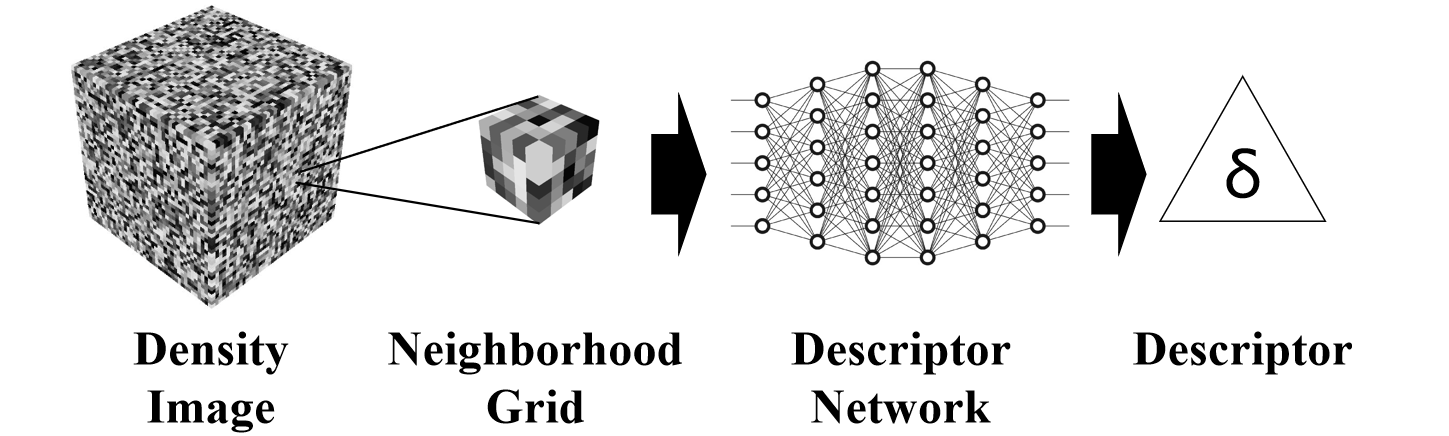}
\end{center}
   \caption{Neural Corner Descriptor Generation. Neighborhood grid is extracted around every corner in every density image.}
\label{fig:2}
\end{figure}

\subsubsection{Universal Descriptor Neural Network}

3D corners as features are more difficult to describe than in 2D, because there exist much more appearance variations of 3D corners than 2D corners. Despite the complexity of 3D corners, we want a descriptor design with simplicity comparable to 2D versions. This leads us to use neural descriptors, in the form of descriptor-generating neural networks (Figure~\ref{fig:2}), which can encode rotation-invariant corner representations within its weights. Given different rotation-dependent corner representations of the same corner as input, the network should output the same result, in order to effectively distinguish corner appearance and orientation.

Previously, we have obtained corner positions in both scenes as indices in multi-scale grids. For each corner, we extract a local neighborhood region of the grid as the input corner representation to the network. For a grid $G$ and its 3D indices of a corner $i,j,k$, we extract a cube-shaped subgrid
\begin{equation}
\label{eq:neighborhood}
N_{i,j,k}(s) = G_{[i-s,i+s]\times[j-s,j+s]\times[k-s,k+s]}
\end{equation}
where $s \in \mathbb{N}^{+}$ indicates the size of the neighborhood. The intervals $[i-s,i+s],[j-s,j+s],[k-s,k+s]$ of grid indices are closed on both sides, so the edge length of $N_{i,j,k}(s)$ is $2s+1$ units. Note that this representation is scale-invariant, because the resolution of $N_{i,j,k}(s)$ is completely determined by the resolution of $G$ and is independent of $s$. Thus we retain scale-invariance after neighborhood extraction and do not rely on the network to be invariant to scale.

The set of $N$ of all corners are then fed into a pre-trained neural network $f$ to generate their descriptors $\delta = f(N)$. This network is universal, in the sense that it can be applied to neighborhood grids sampled from any scene. This is a natural design since corner appearances are pure local geometric features that are independent to global scenes. A network trained from sufficiently diversified corner data is supposed to classify corners in any scene. In addition, our method is flexible in choices of network architectures and output representations $\delta$, which means it can be adapted to multiple descriptor matching criteria. In our experiments, a simple shallow fully-connected network is effective enough for our registration purposes. Please refer to Section~\ref{sec:4} for details.

\subsubsection{Contrastive Learning on Descriptor Networks}
\label{sec:3.3.2}

\begin{figure}[t]
\begin{center}
   \includegraphics[width=1\linewidth]{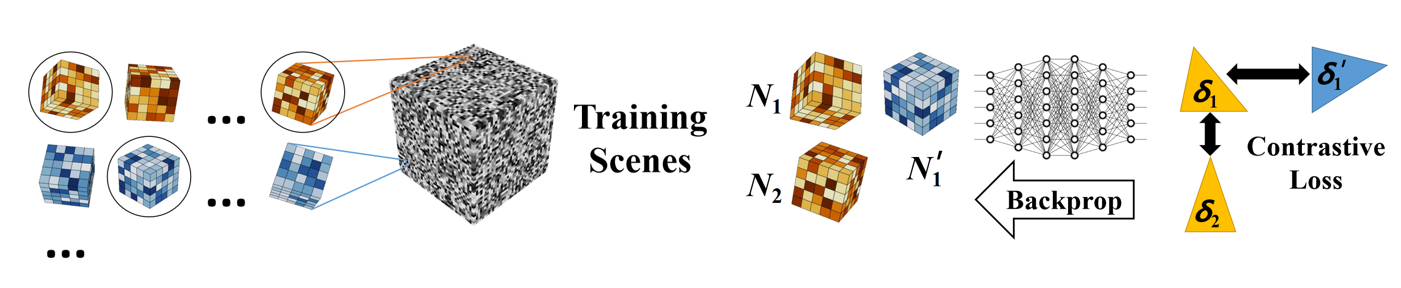}
\end{center}
   \caption{Corner point neighborhoods in various orientations are extracted from training scenes to form triplets, which serves as training data of our descriptor network. The network is trained by a contrastive loss.}
\label{fig:3}
\end{figure}

In this section we describe our method on training the descriptor network, which is a pre-processing step independent of the registration pipeline.  Figure~\ref{fig:3} shows the overall strategy.  Our network is expected to generate similar results, so a contrastive learning strategy is suitable here, where the network is penalized by a contrastive loss that measures the difference between outputs generated from an input pair $(N_1, N_2)$ representing the same corner. However, in addition to matching, we also expect the model to distinguish inputs from corners with different appearances. For this purpose, instead of using training pairs, we choose to use triplets of the form $(N_1, N_2, N^{'}_1)$ to generate $(\delta_1, \delta_2, \delta^{'}_1)$ where $N^{'}_1$ is from a different corner. The contrastive loss penalizes not only the difference between $\delta_1, \delta_2$ but also the similarity between $\delta_1, \delta^{'}_1$. This strategy of learning with triplets has shown to be successful in previous works on 2D image descriptor matching ~\cite{triplet} and object detection ~\cite{FSOD}. We adapt the margin ranking loss first proposed in ~\cite{margin_ranking_loss}:
\begin{equation}
\label{eq:contrastive_loss}
L = \max\{0, \epsilon + ||\delta_1 - \delta_2||_2 - ||\delta_1 - \delta^{'}_1||_2 \}
\end{equation}
where $\epsilon$ is a small positive value, and we assume $\delta$ are vectors with Euclidean difference used. Larger $epsilon$ results in larger penalization on the similarity between $\delta_1, \delta^{'}_1$. However, note that the loss function is flexible as long as it penalizes the aforementioned difference and similarity.

Although there has been no dataset for training corner descriptor networks taking neighborhood grids as input, generating these data can be convenient and effective. We first obtain NeRF models of several scenes, and apply multi-scale corner detection on these scenes as described in Section~\ref{sec:corner_detect}. Then for each corner point, we sample its neighborhood density grid from its original NeRF in various orientations. Since our grids are in 3D space, where there are much more number of orientations for meaningful sampling, we can generate a large number of grids for each detected corner. Each indoor scene we use contains a large number of corners derived from a variety of typical man-made objects, so a small number of scenes is sufficient to generate a sufficiently large and diversified corner dataset for training our network. When training, we randomly sample neighborhood grid triplets from the dataset to compute the contrastive loss and update the weights.

\subsection{Register Key Points Using RANSAC}

After computing descriptors for all corners in both scenes, we match them between the two scenes to get corner point correspondences, and our registration task is converted to traditional point-set registration.

\subsubsection{Matching Descriptors}

Descriptors of the two scenes are matched based on similarity scores $p$, representing the probability for a descriptor pair to be correct. This requires a mathematical definition, which is flexible according to the form of $\delta$. For example, we can use normalized inverse Euclidean distance or angular distance for vector outputs. Descriptor pairs with $p$ smaller than a pre-defined threshold are unlikely to be correct pairs and should be filtered out. The rest of potential matches can be viewed as a bipartite graph, where the vertices consists of descriptors $\delta$, and each edge is associated with a similarity score $p$. Then we apply the Maximum-weight matching algorithm ~\cite{maximum-weight-matching} on this graph to determine descriptor correspondences, which are regarded as correspondences between point location sets $X_1, X_2$ for registration.

\subsubsection{Rigid Registration with RANSAC}

By removing unmatched points from $X_1, X_2$, we finally construct $\Tilde{X}_1, \Tilde{X}_2$ with $|\Tilde{X}_1| = |\Tilde{X}_2|$ for rigid point set registration. For every correct point pair $\mathbf{x}_1 \in \Tilde{X}_1, \mathbf{x}_2 \in \Tilde{X}_2$, we solve for a scale factor $l^{*} > 0$, a $3\times 3$ rotation matrix $R^{*}$ and a 3D translation vector $\mathbf{t}^{*}$ in the rigid transformation that registers $\mathbf{x}_2$ to $\mathbf{x}_1$ with the least registration error of Euclidean distance
\begin{equation}
\label{eq:optimization}
l^{*}, R^{*}, \mathbf{t}^{*} = \argmin_{l, R, \mathbf{t}} \sum_{\substack{(\mathbf{x}_1, \mathbf{x}_2) \\ \mathbf{x}_1 \in \Tilde{X}_1, \mathbf{x}_2 \in \Tilde{X}_2}} n||\mathbf{x}_1 - (lR\mathbf{x}_2 + \mathbf{t})||_2
\end{equation}
where $n = 1$ if pairs are correct and $0$ if otherwise. If all pairs in $\Tilde{X}_1, \Tilde{X}_2$ are perfectly correct, then at least 3 point correspondences are required to determine the transformation. Given such 3 correspondences, there exist algorithms ~\cite{arun, horn} that gives a closed-form solution for $l^{*}, R^{*}$ and $\mathbf{t}^{*}$.

Due to the existence of incorrect correspondences as outliers, we use RANSAC to ignore them. For the transformation $l, R, \mathbf{t}$ proposed by each RANSAC iteration, we consider a point pair $\mathbf{x}_1, \mathbf{x}_2$ to be an inlier if its Euclidean distance error
\begin{equation}
\label{eq:reg_error}
e = ||\mathbf{x}_1 - (lR\mathbf{x}_2 + \mathbf{t})||_2
\end{equation}
is smaller than a pre-defined threshold. Among all transformations with their numbers of inliers larger than a manually set number $m$, we select the one with the least average Euclidean distance error as our final result. $m$ is used to guarantee robust registration results, which is selected according to the performance of the previously used descriptor network. In practice, it is not advised to pick a very small $m$ even if the network performs very well. This is because smaller number of points have higher chance to be symmetric. There may be several transformations between symmetric point correspondences where only one is correct, but the algorithm may not output the correct one.

\section{Experiments}
\label{sec:4}

\begin{figure}
    \begin{minipage}{0.32\textwidth}
        \resizebox{\columnwidth}{!}{\begin{tabular}{llll}
\hline
Layer & Channels & Size & Activation \\
\hline\hline
Input & 1 & N/A & None \\
\hline
Conv3D & 32 & $3\times 3 \times 3$ & ReLU \\
\hline
Conv3D & 64 & $3\times 3 \times 3$ & ReLU \\
\hline
Flatten & 1 & $64 \times 3^3$ & None \\
\hline
Linear & 1 & $2\times 7^3$ & ReLU \\
\hline
Linear & 1 & $7^3$ & None \\
\hline
\end{tabular}}
    \end{minipage}
    \hfill
    \begin{minipage}{0.32\textwidth}
        \includegraphics[width=\textwidth]{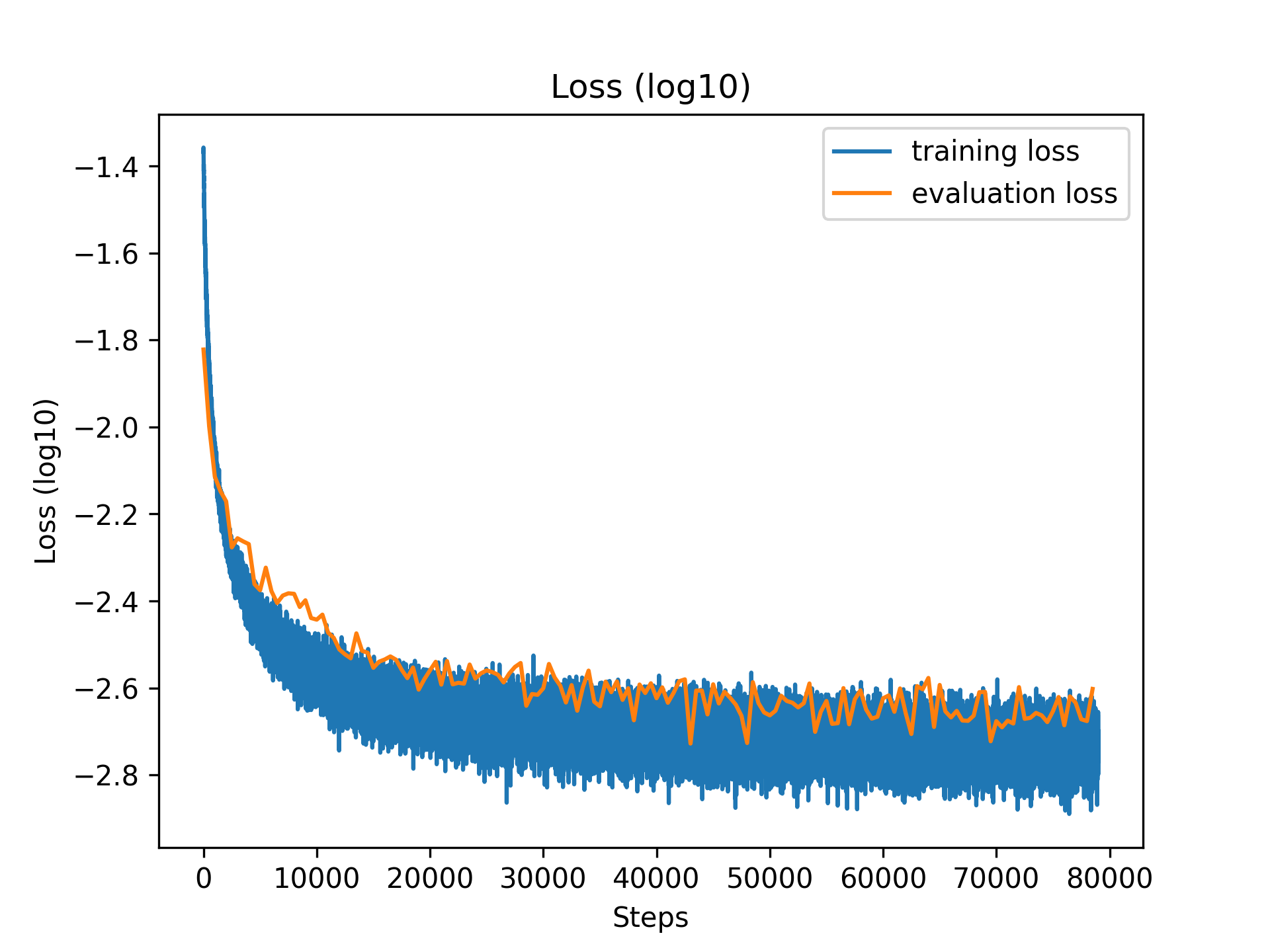}
    \end{minipage}
    \hfill
    \begin{minipage}{0.32\textwidth}
        \includegraphics[width=\textwidth]{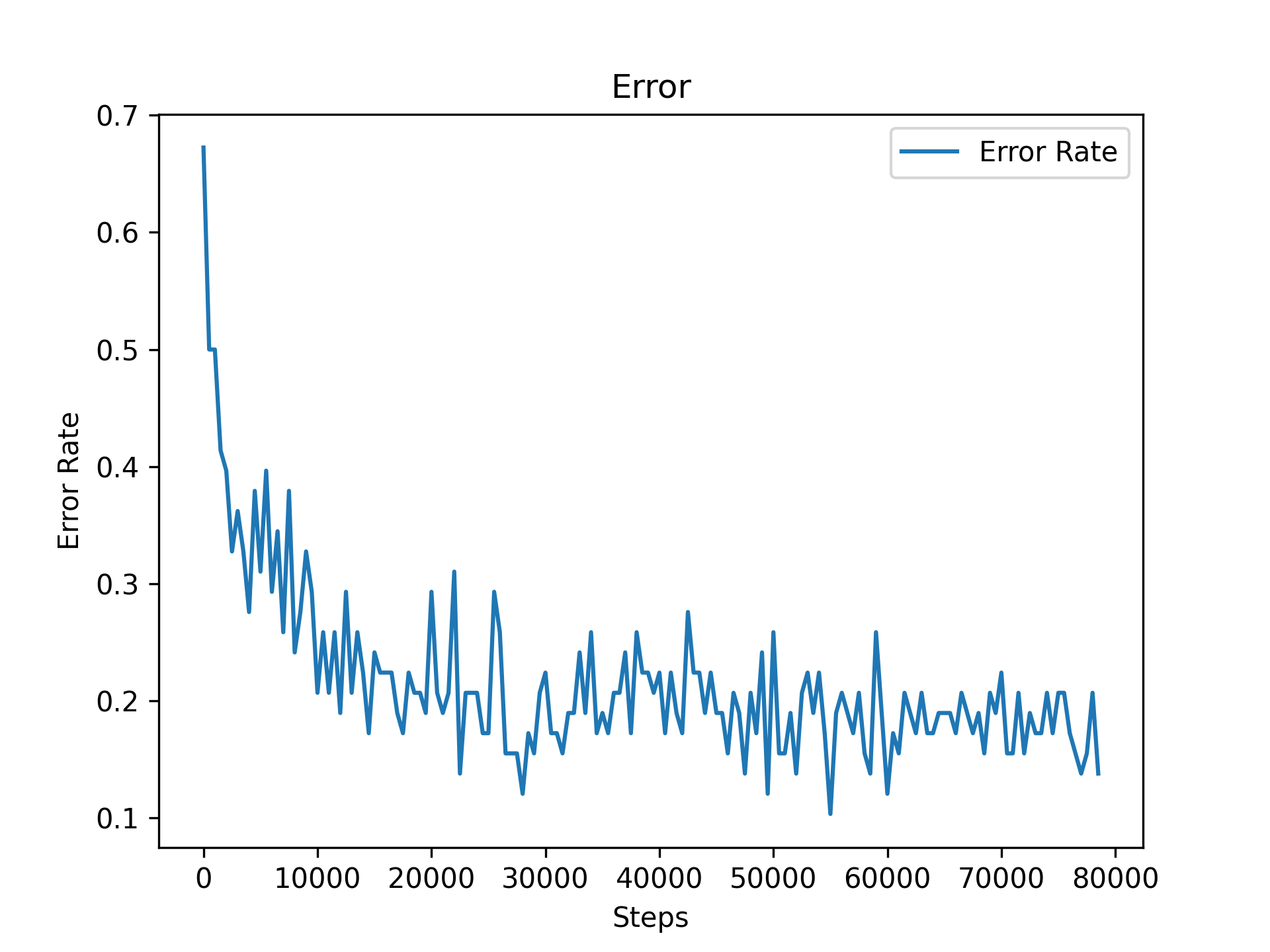}
    \end{minipage}
    \caption{Descriptor network structure (left), training and evaluation loss (center), and error rate (right). Loss is plotted in a log scale. Due to randomness of our triplet formation, loss and error rate have oscillations.}
    \label{fig:4.1}
\end{figure}

\subsection{Training Descriptor Networks}

In our experiments, we extract $7\times7\times7$ neighborhood density grids around corners, feed them into a shallow 3D CNN, and normalize the last layer as the descriptor of the input neighborhood. See Figure~\ref{fig:4.1} for our network structure.

Our network is trained on scenes from the Hypersim~\cite{hypersim} dataset, which is a photorealistic synthetic indoor scene dataset where each scene corresponds with hundreds of rendered images from different viewpoints. In addition to fully rendered images, for each viewpoint, Hypersim also provides images rendered from diffuse illumination only, which we used for training NeRFs with Instant-NGP~\cite{mueller2022instant}. We select 22 scenes providing about 1200 corners for training our network and 1 scene providing 58 corners for validation.

We implement Section~\ref{sec:3.3.2} to generate neighborhood grids from these scenes. For each scene, we select corners from detected corners not on the scene boundary. For each detected corner, its $7\times 7\times 7$ neighborhood grid is rotated along the $x,y,z$ axes by $\theta_x, \theta_y, \theta_z$, where $\theta_x, \theta_y, \theta_z$ take values from evenly-spaced angles in $(-\pi, \pi ] $. Each angle is spaced by $\frac{1}{6}\pi$, so for each corner, we generate $(\frac{2\pi}{\frac{1}{6}\pi})^3 = 3456$ neighborhood density grids. These grids derived from different corners from different scenes are then assembled as our training data. To form a training triplet, we randomly select 2 grids from one corner and 1 grid from another corner. In every training iteration, 10,000 triplets are formed and fed into the network to compute the contrastive loss defined in Equation~\ref{eq:contrastive_loss}, where we use $\epsilon = 0.1$.  The Adam~\cite{adam} optimizer with learning rate $2\times 10^{-6}$ is used. We train our network on an NVIDIA GeForce GTX 1080 Ti GPU for 80,000 iterations, which takes about 7 hours.

The training and validation loss are shown in Figure~\ref{fig:4.1}. Due to randomness of our triplet formation, the two loss curves exhibit some oscillations. In addition to loss, we also compute error rate on validation data for a more direct evaluation on the effectiveness of corner classification. In each validation step, we randomly select 1 test neighborhood grid from each corner. For each test grid, another 58 grids from these 58 corners are randomly selected as proposal grids to compute similarity scores with the test grid. The one with the largest similarity score, for which we use inverse angular distance, is matched to the test grid. This match is correct if its two grids come from the same corner. Then, the error rate is computed by $n_{\text{wrong}}/100$ where $n_{\text{wrong}}$ is the number of incorrect matches. Please refer to Figure~\ref{fig:4.1} for error rates as training progresses. In the end, error rates remain around 0.2. According to RANSAC, given our network performance, roughly 7 corner pairs from the overlapping region are required for a robust registration result with 99\% confidence. This is not a very strict requirement which is often easily satisfied.

\begin{figure*}\centering
     \begin{subfigure}[b]{0.195\textwidth}
         \centering
         \includegraphics[width=\textwidth]{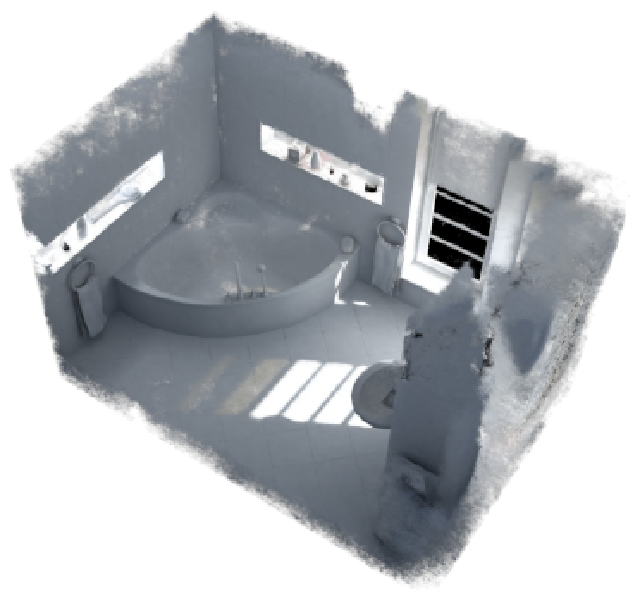}
     \end{subfigure}
     \hfill
     \begin{subfigure}[b]{0.195\textwidth}
         \centering
         \includegraphics[width=\textwidth]{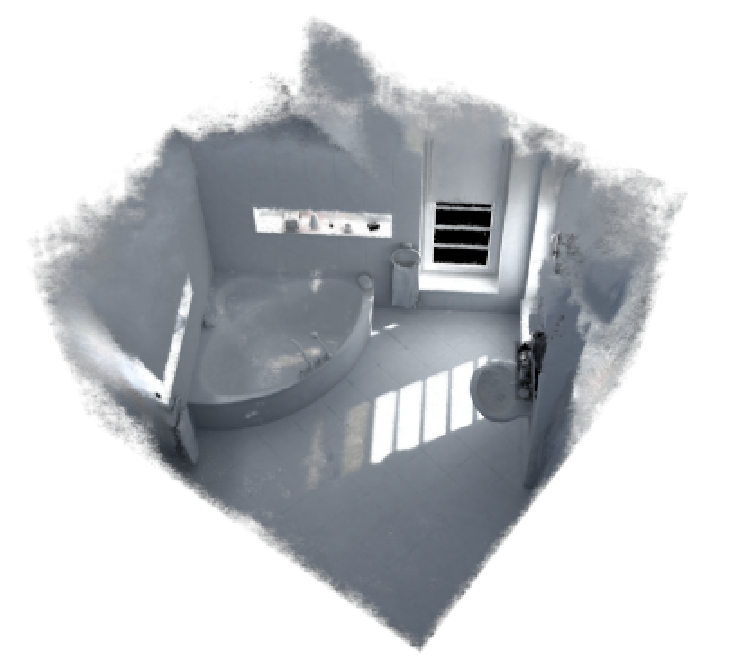}
     \end{subfigure}
     \hfill
     \begin{subfigure}[b]{0.195\textwidth}
         \centering
         \includegraphics[width=\textwidth]{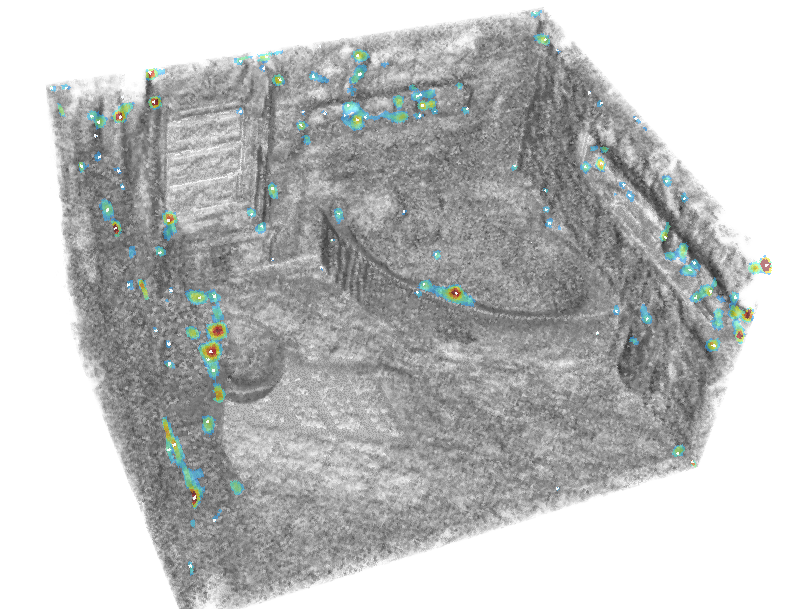}
     \end{subfigure}
     \hfill
     \begin{subfigure}[b]{0.195\textwidth}
         \centering
         \includegraphics[width=\textwidth]{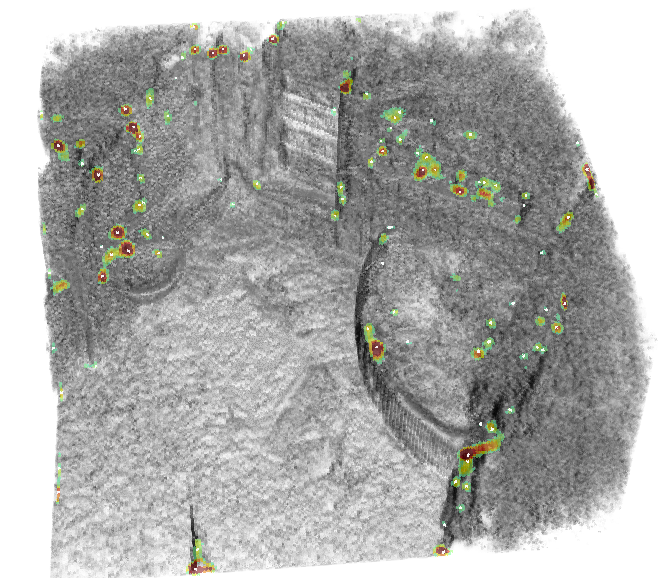}
     \end{subfigure}
     \hfill
     \begin{subfigure}[b]{0.195\textwidth}
         \centering
         \includegraphics[width=\textwidth]{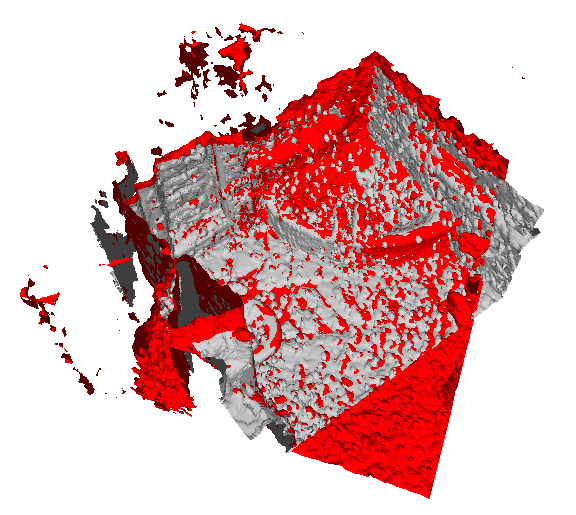}
     \end{subfigure}
     \hfill
     \begin{subfigure}[b]{0.195\textwidth}
         \centering
         \includegraphics[width=\textwidth]{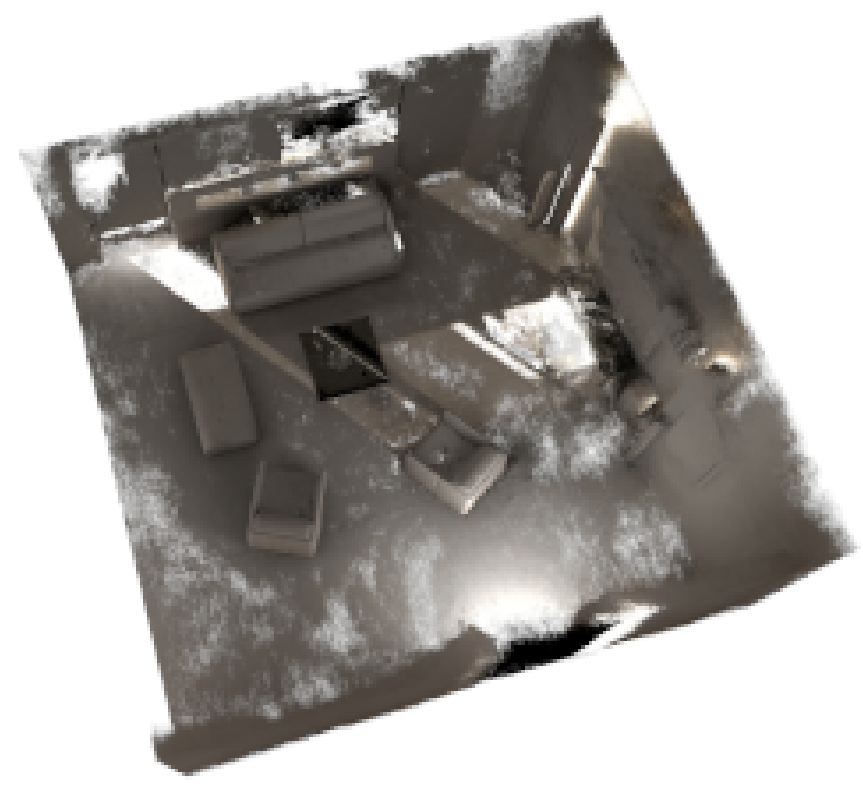}
     \end{subfigure}
     \hfill
     \begin{subfigure}[b]{0.195\textwidth}
         \centering
         \includegraphics[width=\textwidth]{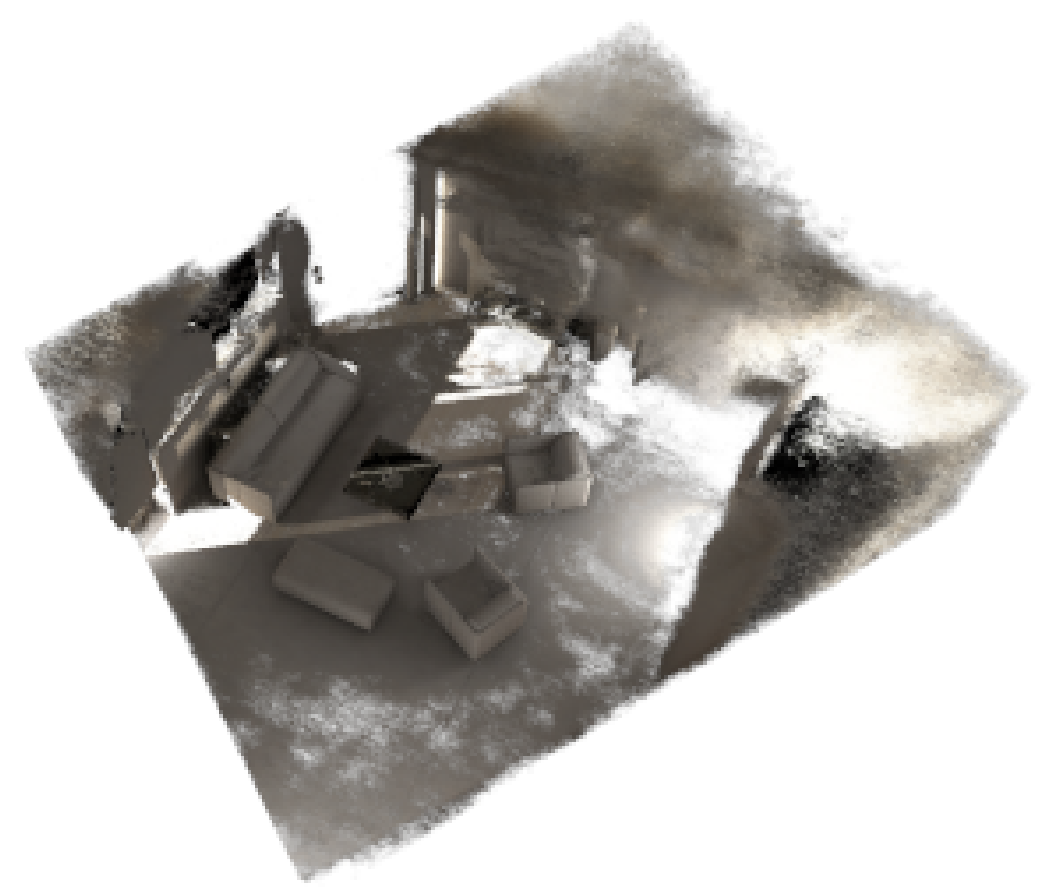}
     \end{subfigure}
     \hfill
     \begin{subfigure}[b]{0.195\textwidth}
         \centering
         \includegraphics[width=\textwidth]{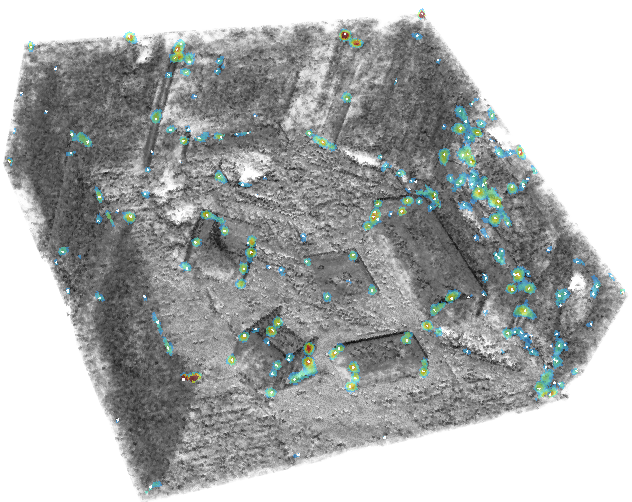}
     \end{subfigure}
     \hfill
     \begin{subfigure}[b]{0.195\textwidth}
         \centering
         \includegraphics[width=\textwidth]{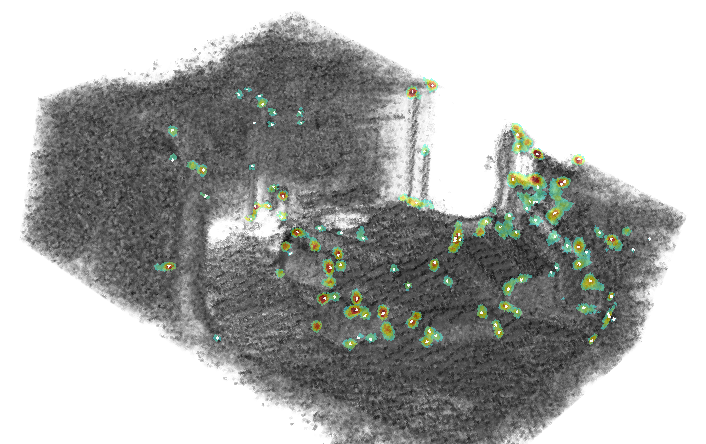}
     \end{subfigure}
     \hfill
     \begin{subfigure}[b]{0.195\textwidth}
         \centering
         \includegraphics[width=\textwidth]{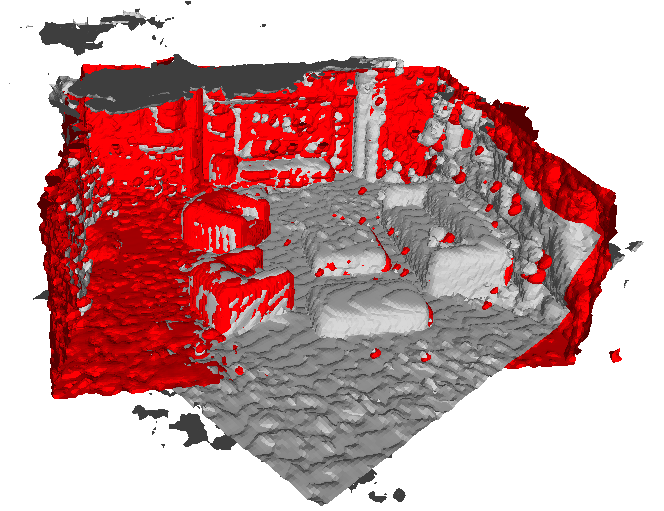}
     \end{subfigure}
     \hfill
     \begin{subfigure}[b]{0.195\textwidth}
         \centering
         \includegraphics[width=\textwidth]{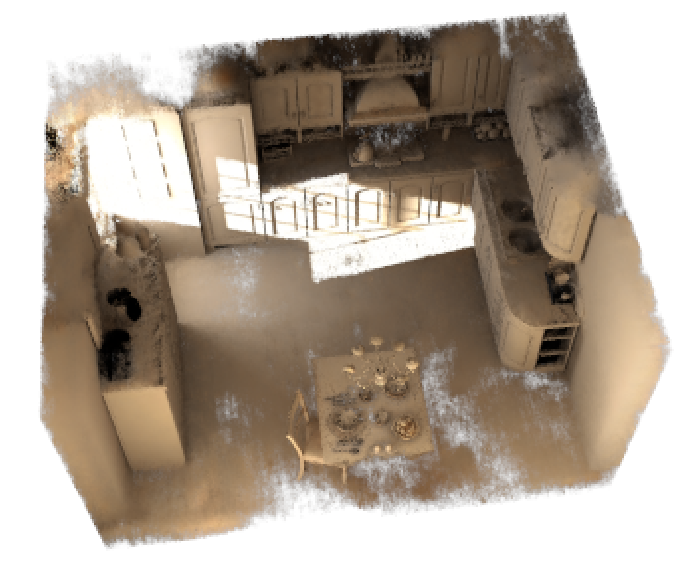}
     \caption{NeRF 1}
     \end{subfigure}
     \hfill
     \begin{subfigure}[b]{0.195\textwidth}
         \centering
         \includegraphics[width=\textwidth]{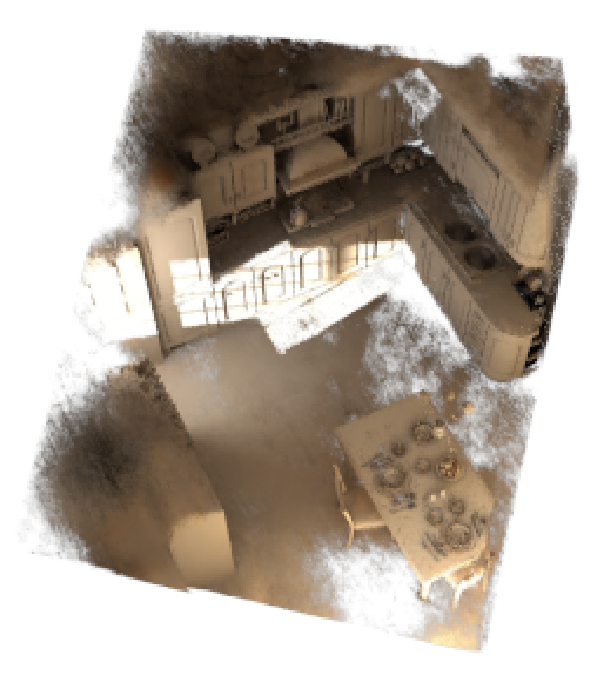}
     \caption{NeRF 2}
     \end{subfigure}
     \hfill
     \begin{subfigure}[b]{0.195\textwidth}
         \centering
         \includegraphics[width=\textwidth]{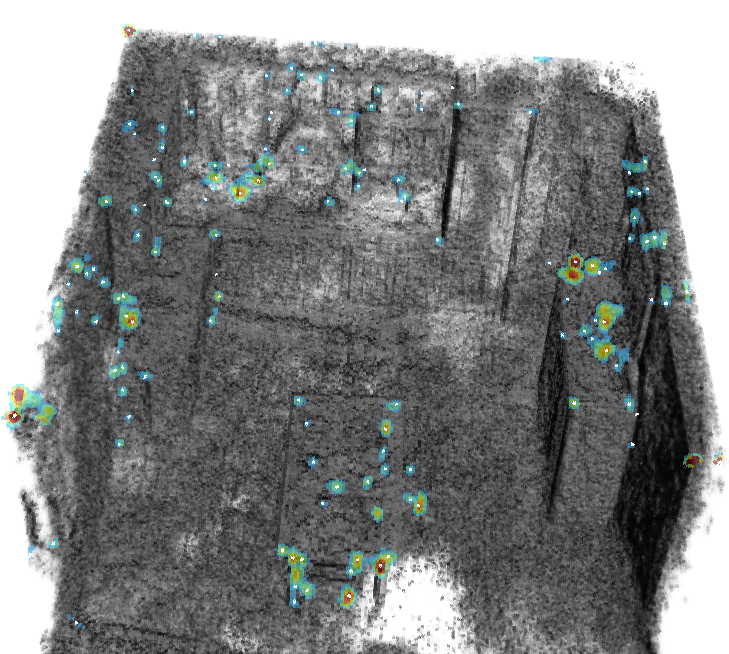}
     \caption{Harris 1}
     \end{subfigure}
     \hfill
     \begin{subfigure}[b]{0.195\textwidth}
         \centering
         \includegraphics[width=\textwidth]{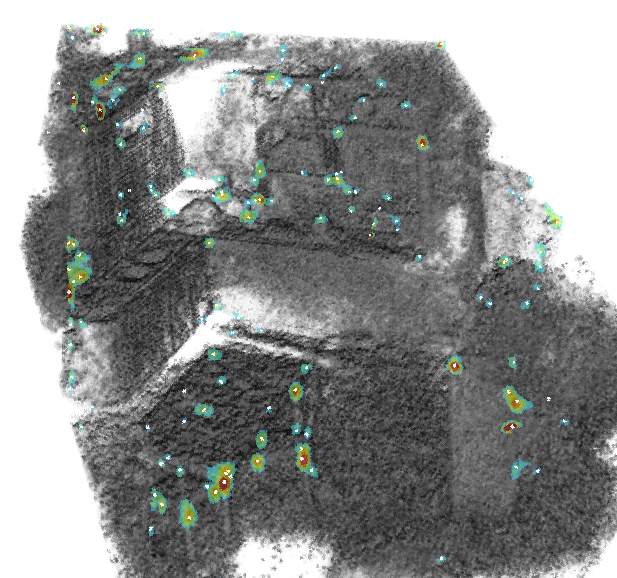}
     \caption{Harris 2}
     \end{subfigure}
     \hfill
     \begin{subfigure}[b]{0.195\textwidth}
         \centering
         \includegraphics[width=\textwidth]{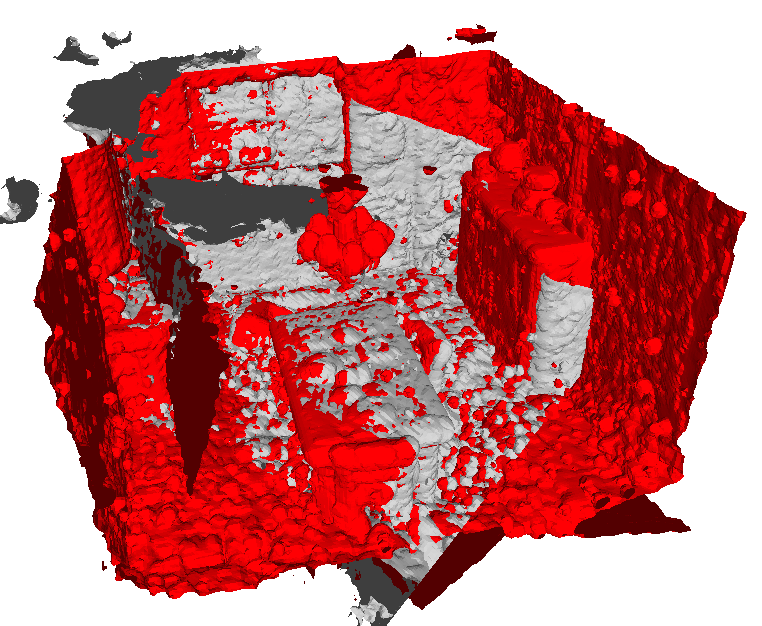}
     \caption{Results}
     \end{subfigure}
     \hfill
\caption{Visualization of the NeRFs, density grids with Harris corners as white dots and response heatmap around them, and registered results. The first, second, third row corresponds to Hypersim scene ai\_001\_001, ai\_001\_008, ai\_002\_005 respectively. Each row visualizes two parts of the scene to be registered, as well as the registered volume in the last column rendered as two meshes with different colors.}
\label{fig:4.2}
\end{figure*}

\begin{figure}
    \begin{minipage}{0.49\textwidth}
\centering
        \resizebox{\columnwidth}{!}{\begin{tabular}{|l|l|l|l}
\hline
Net\textbackslash Sift & Success & Fail \\
\hline
Success & 9 & 4 \\
\hline
Fail & 1 & 1 \\
\hline
\end{tabular}}
\end{minipage}
    \begin{minipage}{0.49\textwidth}
\centering
        \resizebox{\columnwidth}{!}{\begin{tabular}{|l|l|}
\hline
Descriptor & Average Error \\
\hline
Net & 0.839 \\
\hline
Sift & 0.822 \\
\hline
\end{tabular}}
\end{minipage}
    \caption{Count of successes and failures of both types of descriptors (left); and the average registration error on the successful cases of both descriptors (right). `Net' denotes our descriptor network, and `Sift' denotes 3D-SIFT. Note that although Network-based and sift-based descriptors give similar registration error on successful cases, descriptor network has more number of successes.}
    \label{table:4.2}
\end{figure}

\subsection{Registration Results}
\label{sec:4.2}

We use the network described and trained above to measure the performance of our method. The Hypersim dataset is used for training and testing as well as validation to ensure network coherency. We note on the other hand the Hypersim dataset has its inherent deficiencies. Specifically, camera poses are not abundant enough for a single scene, which leads to information loss in the opposite direction. In many cases, only the front view of the room is visible and thus is not suitable for registration. To avoid such corruption of data, we picked 15 scenes of relatively high quality and not used in training, which are then cropped into overlapping scene portions from different orientations for registration.

For each scene, we manually split its NeRF into 2 parts with partial overlapping, and sample density grids in different resolutions. We translate and rotate one of the density grid and try to register back to see if the two volumes register as well as the registration error. We run RANSAC for 50,000 iterations, where a transformation is only considered if the number of inlier pairs is larger than 6. A point pair is considered an inlier if their registered distance error is less than $3$ (we regard a grid cell unit length to be 1). The qualitative visualization of registration on 3 scenes are shown in Figure~\ref{fig:4.2}. On the other hand, we measure the registration results quantitatively on 15 Hypersim scenes by the average squared distance error. This error is defined as
\begin{equation}
\label{eq:avg_error}
e_{\text{avg}} = \frac{1}{|I|} \sum_{(\mathbf{x}_1, \mathbf{x}_2) \in I} ||\mathbf{x}_1 - (lR\mathbf{x}_2 + \mathbf{t})||_2^2
\end{equation}
where $I$ is the set of inlier corner point pairs determined by RANSAC. These error are summarized in Figure~\ref{table:4.2}. Typically, the magnitude of distance errors is comparable to the grid cell length, which indicate very good registration results for large indoor scenes.

\subsection{Comparison with 3D-SIFT}

While we use neural networks to learn 3D corner descriptors, it is natural to question about the performance of traditional, non-learning-based or hand-crafted descriptors. In this section, we compare our neural descriptor with a typical traditional descriptor 3D SIFT~\cite{3dsift}. 3D SIFT computes circular histograms of gradient orientations in subgrids of the local neighborhood grid. The histograms are concatenated together, normalized and rotated to align to the dominant gradient direction, thus it is rotation-invariant.

In our experiments, we extract $9\times 9 \times 9$ neighborhoods around each detected corner. This neighborhood is evenly divided into 27 subgrids to compute histograms for each of them. Then we replace our neural descriptors with these histograms in our pipeline and perform registration on the test scenes used in Section~\ref{sec:4.2}. Other settings are identical to our previous experiments on the network-based descriptor. The results are shown in Figure~\ref{table:4.2}.

As shown in this table, network-based registration has larger numbers of successful registration attempts than 3D-SIFT. In addition, once a descriptor netowrk is loaded, it can generate corner descriptors very efficiently, while 3D-SIFT has to take about half a minute to compute the descriptor. This shows that our descriptor network performs better than 3D-SIFT, so we believe it is a better descriptor choice for our 3D-image based NeRF registration.

\begin{figure*}\centering
     \begin{subfigure}[b]{0.24\textwidth}
         \centering
         \includegraphics[width=\textwidth]{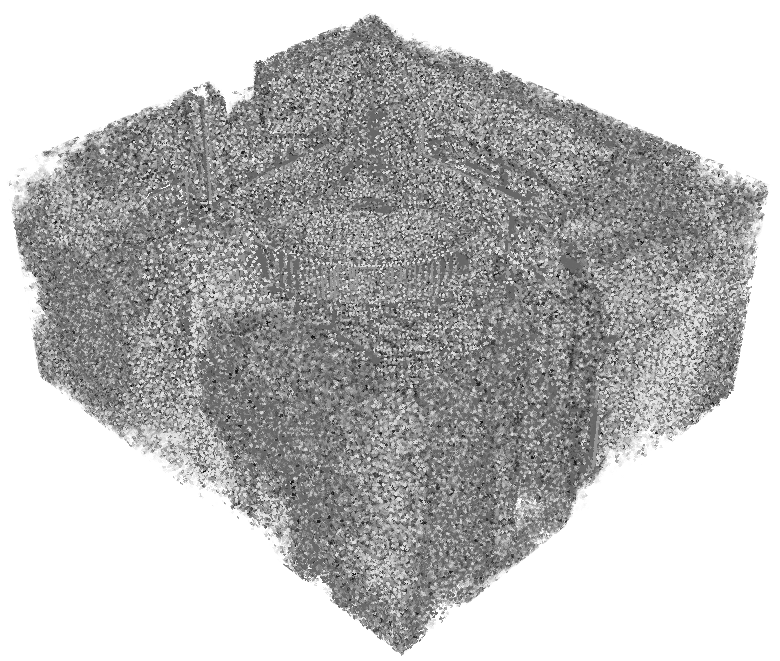}
     \end{subfigure}
     \hfill
     \begin{subfigure}[b]{0.24\textwidth}
         \centering
         \includegraphics[width=\textwidth]{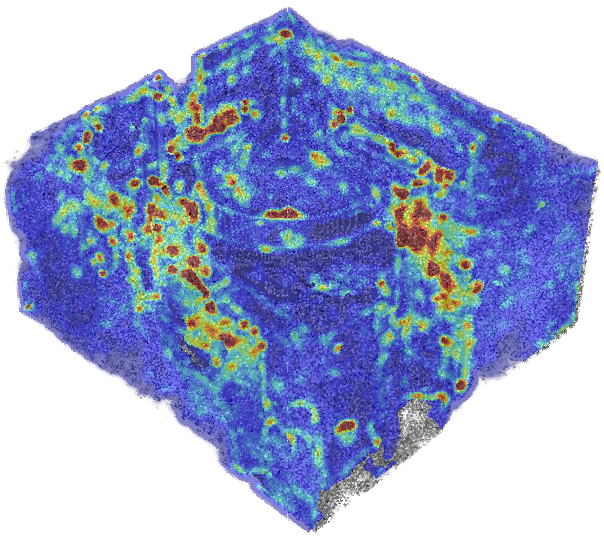}
     \end{subfigure}
     \hfill
     \begin{subfigure}[b]{0.24\textwidth}
         \centering
         \includegraphics[width=\textwidth]{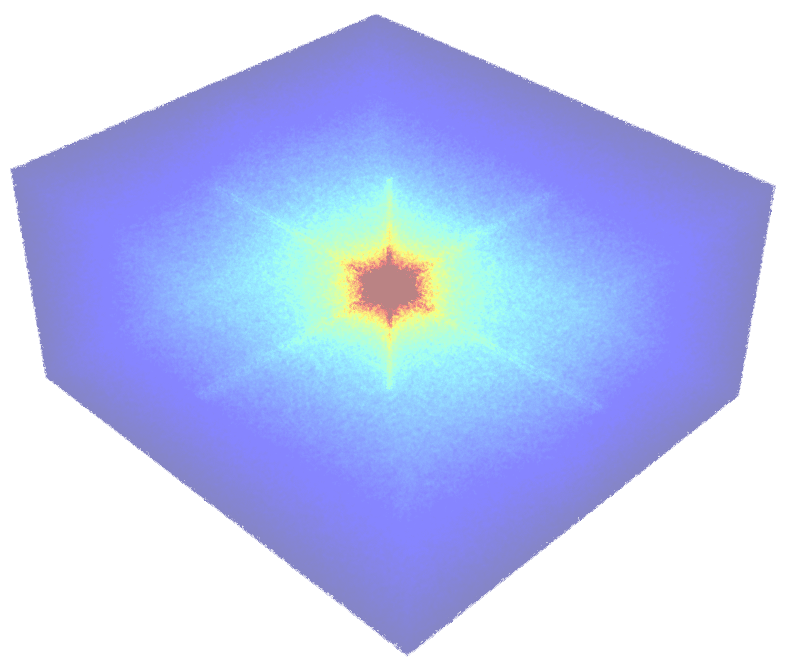}
     \end{subfigure}
     \hfill
     \begin{subfigure}[b]{0.24\textwidth}
         \centering
         \includegraphics[width=\textwidth]{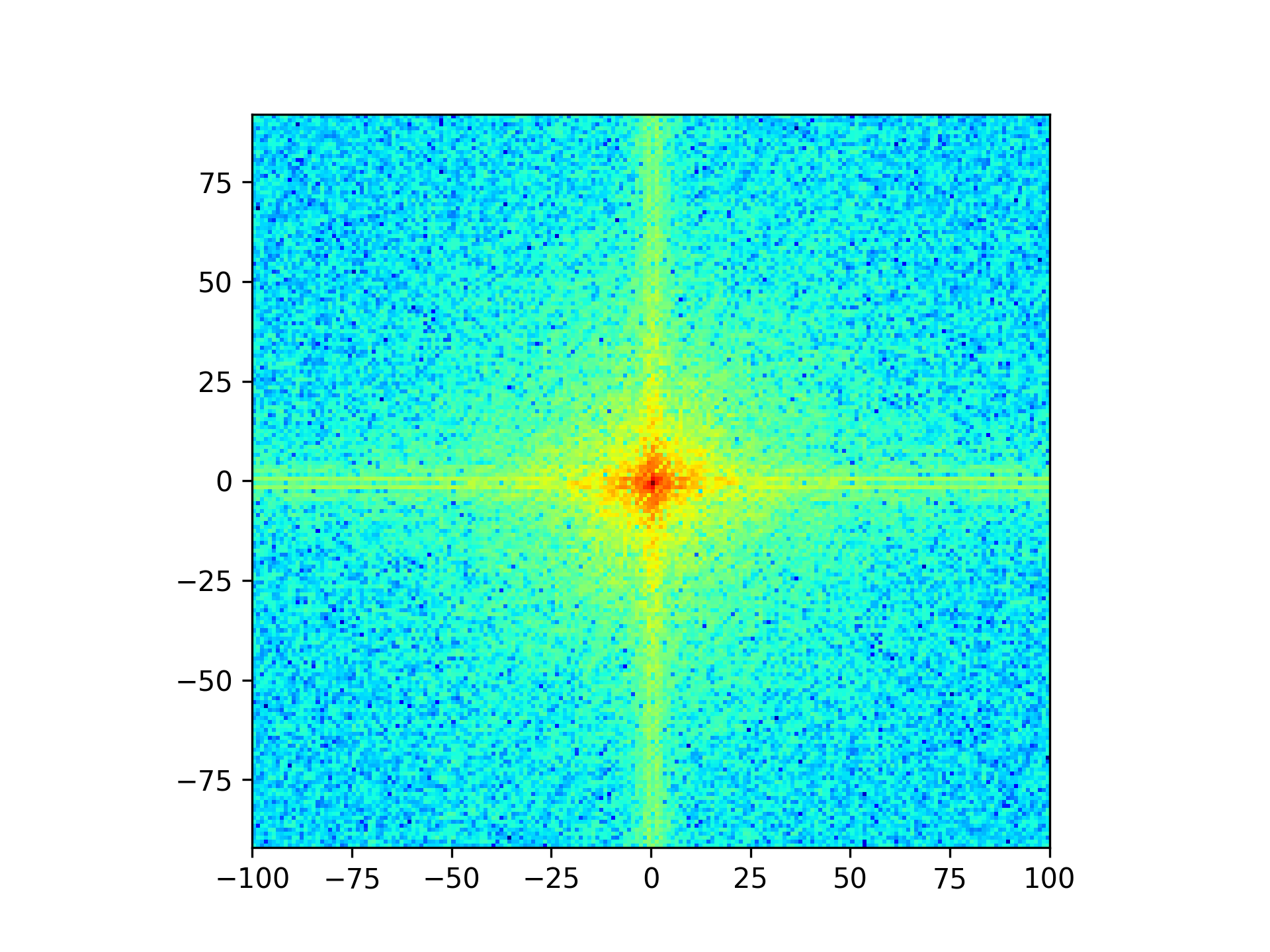}
     \end{subfigure}
     \hfill
     \begin{subfigure}[b]{0.24\textwidth}
         \centering
         \includegraphics[width=\textwidth]{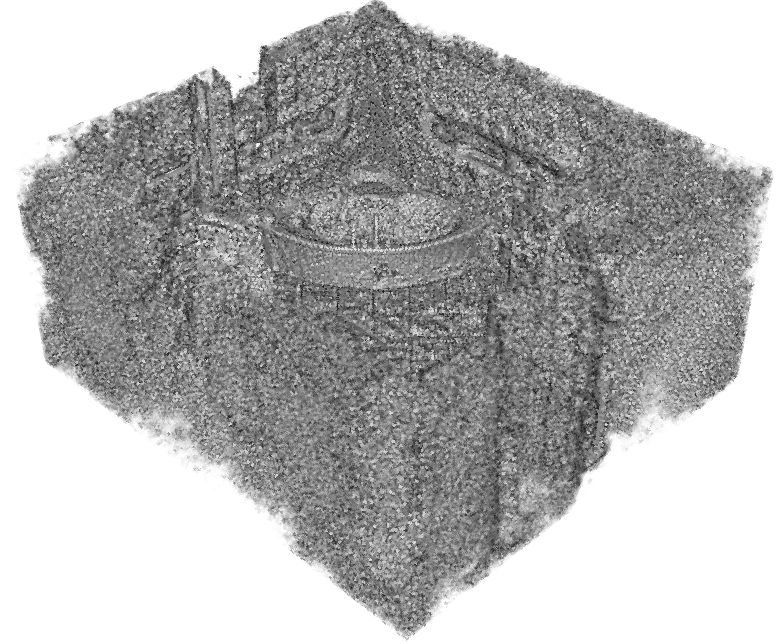}
     \end{subfigure}
     \hfill
     \begin{subfigure}[b]{0.24\textwidth}
         \centering
         \includegraphics[width=\textwidth]{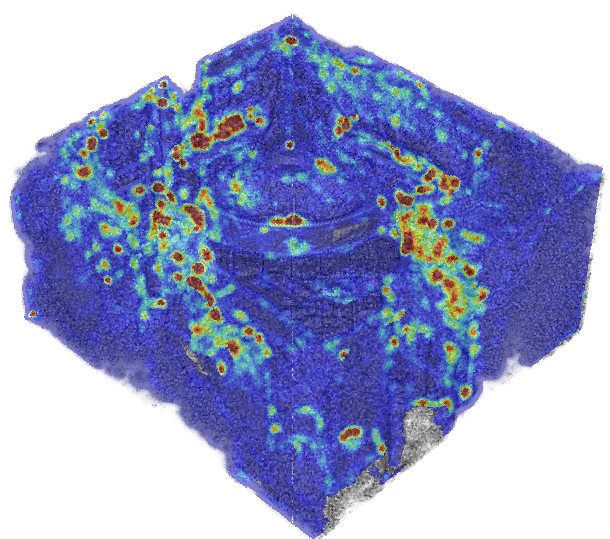}
     \end{subfigure}
     \hfill
     \begin{subfigure}[b]{0.24\textwidth}
         \centering
         \includegraphics[width=\textwidth]{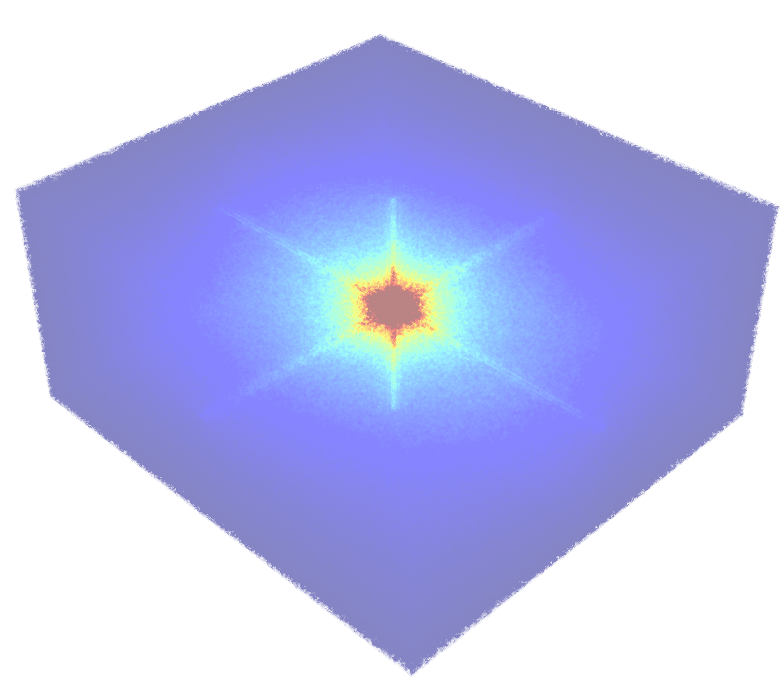}
     \end{subfigure}
     \hfill
     \begin{subfigure}[b]{0.24\textwidth}
         \centering
         \includegraphics[width=\textwidth]{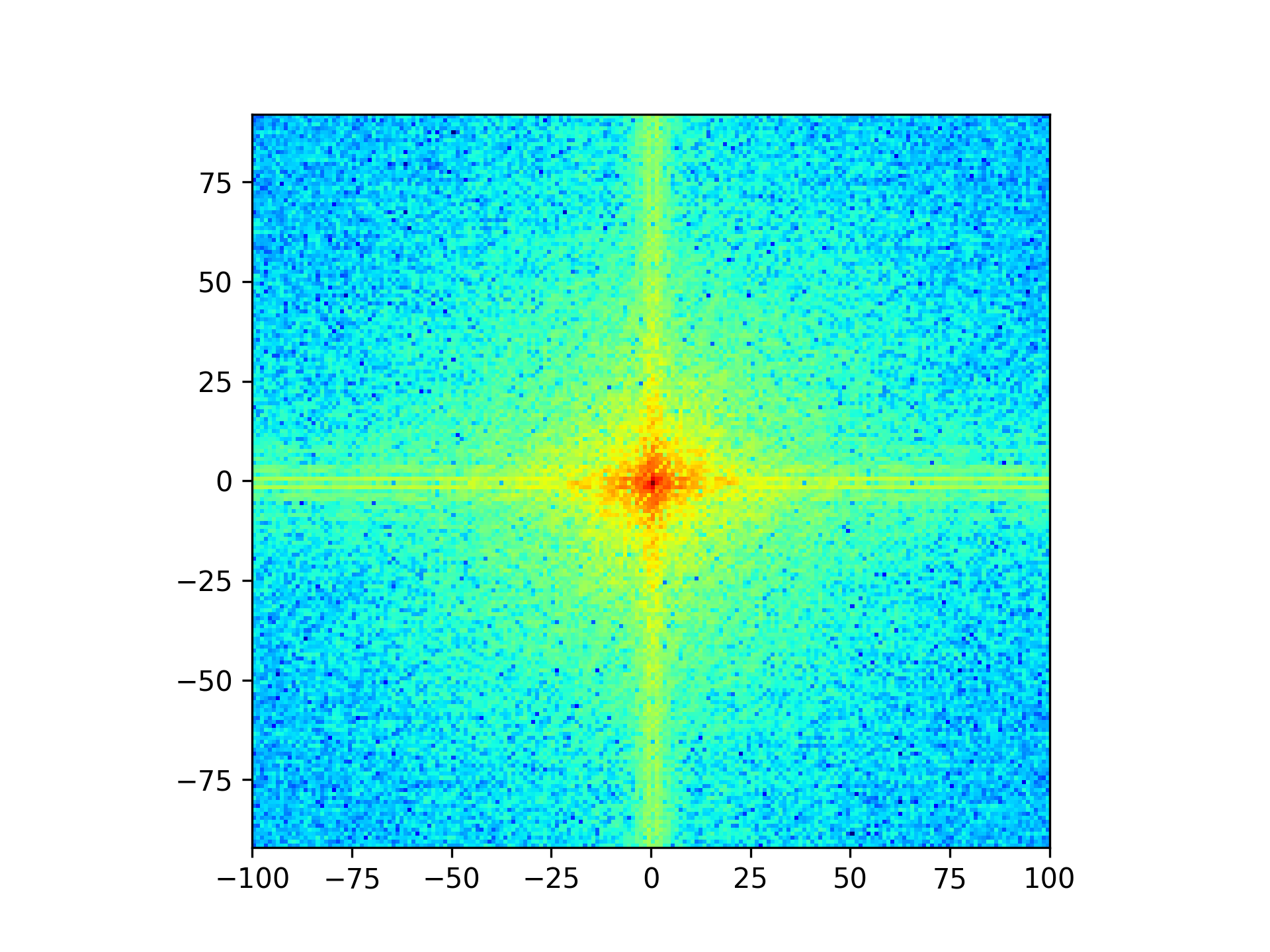}
     \end{subfigure}
     \hfill
     \begin{subfigure}[b]{0.24\textwidth}
         \centering
         \includegraphics[width=\textwidth]{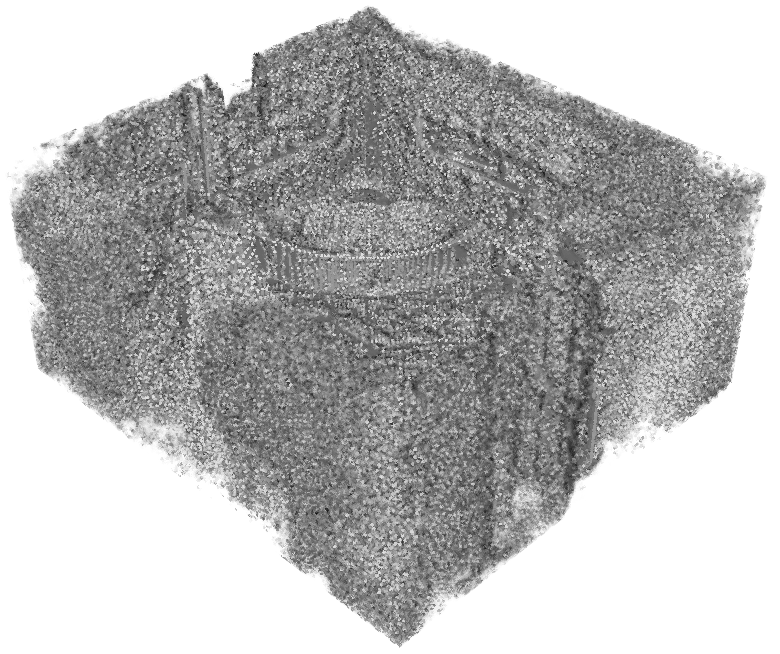}
     \caption{Density Grid}
     \end{subfigure}
     \hfill
     \begin{subfigure}[b]{0.24\textwidth}
         \centering
         \includegraphics[width=\textwidth]{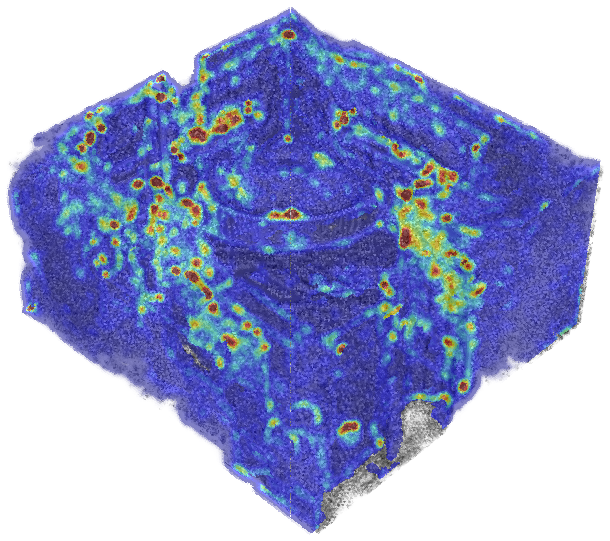}
     \caption{Harris Heatmap}
     \end{subfigure}
     \hfill
     \begin{subfigure}[b]{0.24\textwidth}
         \centering
         \includegraphics[width=\textwidth]{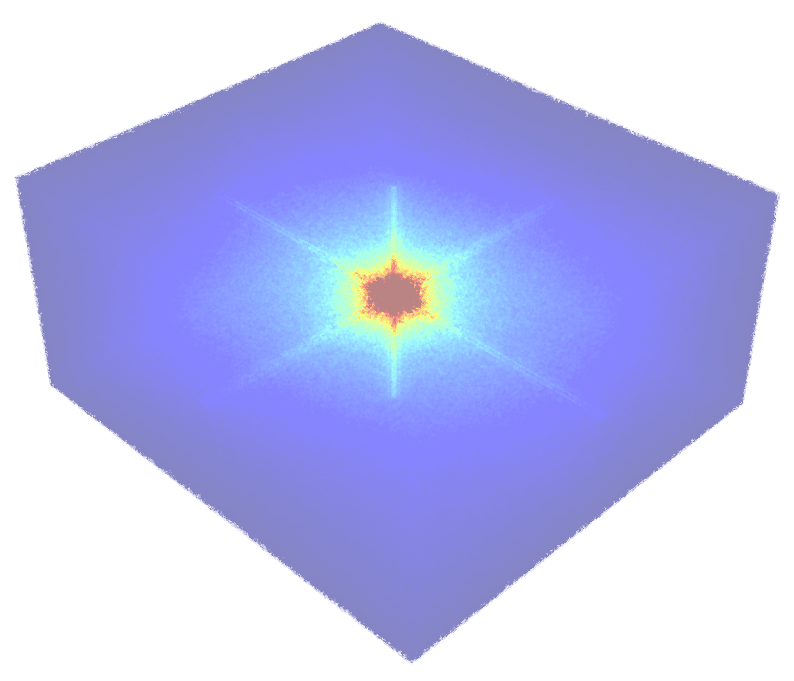}
     \caption{3D Frequency}
     \end{subfigure}
     \hfill
     \begin{subfigure}[b]{0.24\textwidth}
         \centering
         \includegraphics[width=\textwidth]{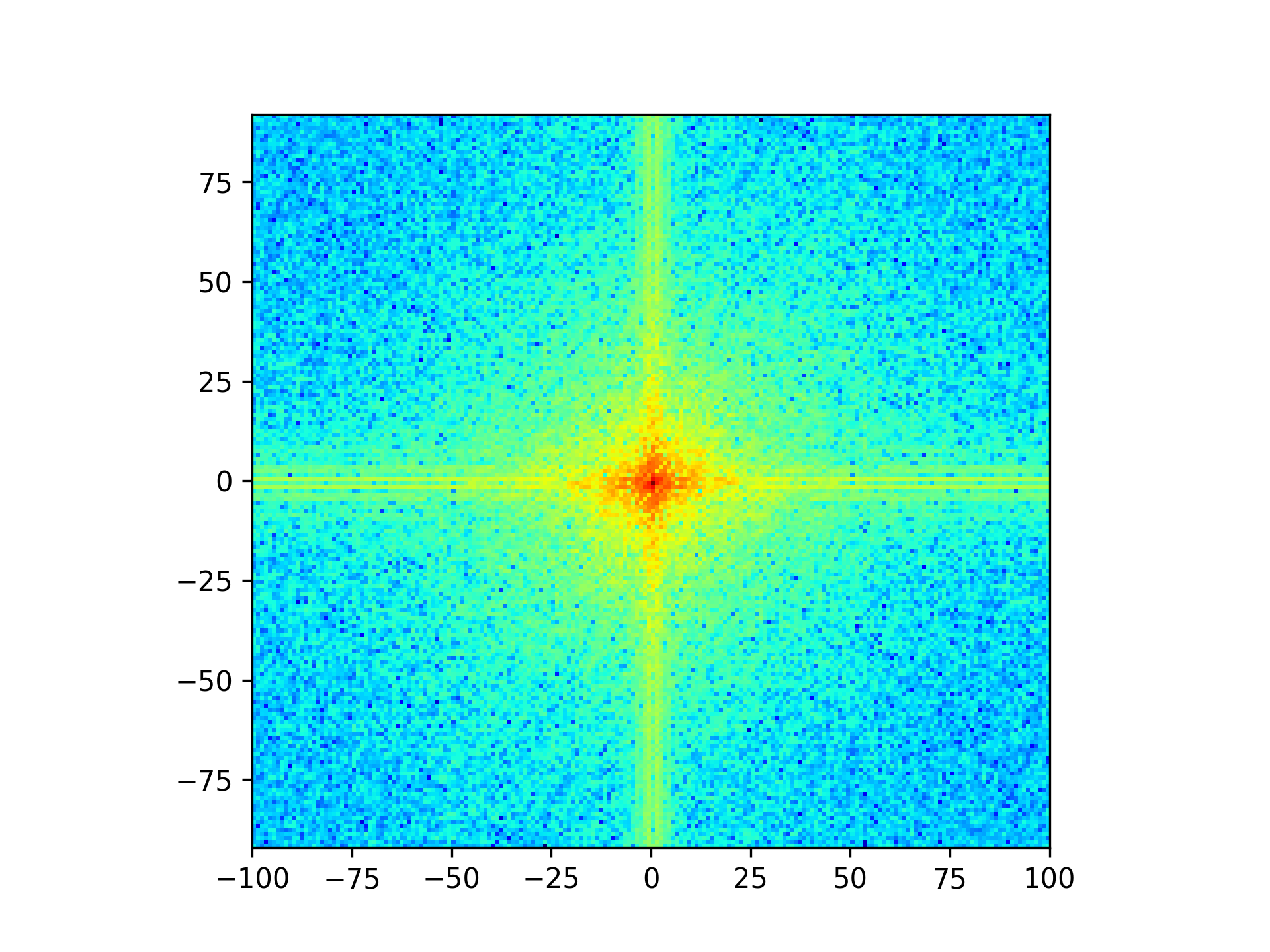}
     \caption{2D Frequency}
     \end{subfigure}
     \hfill
\caption{Visualization of density grid, Harris response and frequency domains. Row 1: Direct sampling. Row 2: Average Pooling. Row 3: Anisotropic Diffusion. The `jet' color map is used where red denotes higher value and blue denotes smaller value. All grids and images have values normalized within 0 and 1. For the frequency domains, zero-frequency locate at the center, with higher positive and negative frequencies around.}
\label{fig:4.4}
\end{figure*}

\subsection{Neural Density Field Sampling Strategies}
\label{sec:4.4}

Here we compare 3 different density field sampling strategies mentioned in section~\ref{sec:3.1}, namely direct sampling, average pooling from higher resolution, and anisotropic diffusion. For average pooling, we use pool size $2\times 2\times 2$ and apply it once. For anisotropic diffusion, we use 5 iterations with timestep $\Delta t = 0.01$. The diffusion coefficient $c$ we use is
\begin{equation}
\label{eq:aniso}
c(||\nabla I||) = e^{-(\frac{||\nabla I||}{K})^2}
\end{equation}
where $\nabla I$ denotes the Laplacian of the grid, and the sensitivity constant $K = 5$. We select a typical scene from Hypersim and sample with each strategy. For all 3 strategies, we visualize their resulting density grid, normalized Harris response and frequency domain of density in Figure~\ref{fig:4.4}. For the frequency domain, we visualize the amplitudes of frequencies obtained by 3D Discrete Fourier Transform. For better illustration, we also visualize 2D frequency domains by slicing 3D domains at zero-frequency on an axis.

Density grid directly sampled from NeRF appears to be noisy, which also causes its Harris response to be noisy. Correspondingly, we can see high amplitudes for higher frequencies in its frequency domain. By contrast, the other grids are less noisy, and the high amplitudes in their frequency domains centralize around the lower frequencies. For denoising, as seen in the frequency domains of average pooling and anisotropic diffusion, their effectiveness is similar. However, the density grid obtained from average pooling looks blurrier than that from anisotropic diffusion. Average pooling also causes higher normalized Harris response around corners, showing its smoothing effects on corners. To conclude, anisotropic diffusion does the best in denoising and corner-preserving, so it is applied in our sampling step.

\section{Limitation}
\label{sec:5}

As a method that generalizes from key point based 2D image registration, our framework has several limitations. First, the accuracy of registration is limited by the resolution of the sampling density grids, which means that optimizing registration in the scale of a grid cell unit length is not possible. This may require a method that directly operates on the continuous field. In addition, our method relies on a sufficient number of correct matches between corner points. If the overlapping part of 2 NeRFs does not contain enough corners as key points, our framework is expected to struggle.

\section{Conclusion}
\label{sec:6}

We propose a NeRF registration framework which operates on a 3D image representation of the NeRF density field. This framework generalizes the traditional 2D image registration pipeline to 3D. We propose to use a universal descriptor network to generate descriptors of 3D corner features without fine-tuning, as well as a contrastive learning strategy and a data generation method for training the network. By performing experiments on the Hypersim dataset, we demonstrate that our framework can register two indoor scenes with sufficient accuracy. We also show that the performance of our shallow fully-connected descriptor network is adequate for our registration purpose. As the first effort on direct NeRF registration, we hope that our framework can benefit the construction of NeRFs of large-scale indoor scenes, and inspire future work on NeRF registration.

{
\small
\bibliographystyle{ieee_fullname}
\bibliography{ref}

\begin{thebibliography}{10}\itemsep=-1pt

\bibitem{abdel2006csift}
Alaa~E Abdel-Hakim and Aly~A Farag.
\newblock Csift: A sift descriptor with color invariant characteristics.
\newblock In {\em 2006 IEEE computer society conference on computer vision and
  pattern recognition (CVPR'06)}, volume~2, pages 1978--1983. Ieee, 2006.

\bibitem{allaire2008full}
St{\'e}phane Allaire, John~J Kim, Stephen~L Breen, David~A Jaffray, and
  Vladimir Pekar.
\newblock Full orientation invariance and improved feature selectivity of 3d
  sift with application to medical image analysis.
\newblock In {\em 2008 IEEE computer society conference on computer vision and
  pattern recognition workshops}, pages 1--8. IEEE, 2008.

\bibitem{arun}
K.~S. Arun, T.~S. Huang, and S.~D. Blostein.
\newblock Least-squares fitting of two 3-d point sets.
\newblock {\em IEEE Transactions on Pattern Analysis and Machine Intelligence},
  PAMI-9(5):698--700, 1987.

\bibitem{bay2006surf}
Herbert Bay, Tinne Tuytelaars, and Luc Van~Gool.
\newblock Surf: Speeded up robust features.
\newblock {\em Lecture notes in computer science}, 3951:404--417, 2006.

\bibitem{NeRF_2020}
Matthew Tancik Jonathan T. Barron Ravi Ramamoorthi Ren~Ng Ben~Mildenhall,
  Pratul P.~Srinivasan.
\newblock Nerf: Representing scenes as neural radiance fields for view
  synthesis.
\newblock In {\em European Conference on Computer Vision (ECCV)}, 2020.

\bibitem{brown2005multi}
Matthew Brown, Richard Szeliski, and Simon Winder.
\newblock Multi-image matching using multi-scale oriented patches.
\newblock In {\em 2005 IEEE Computer Society Conference on Computer Vision and
  Pattern Recognition (CVPR'05)}, volume~1, pages 510--517. IEEE, 2005.

\bibitem{multires_mesh}
XueSong Chen, LiangJun Zhang, RuoFeng Tong, and JinXiang Dong.
\newblock Multi-resolution-based mesh registration.
\newblock In {\em 8th International Conference on Computer Supported
  Cooperative Work in Design}, volume~1, pages 88--93 Vol.1, 2004.

\bibitem{chen2022local}
Yue Chen, Xingyu Chen, Xuan Wang, Qi Zhang, Yu Guo, Ying Shan, and Fei Wang.
\newblock Local-to-global registration for bundle-adjusting neural radiance
  fields.
\newblock {\em arXiv preprint arXiv:2211.11505}, 2022.

\bibitem{nsift}
Warren Cheung and Ghassan Hamarneh.
\newblock N-sift: N-dimensional scale invariant feature transform for matching
  medical images.
\newblock In {\em 2007 4th IEEE international symposium on biomedical imaging:
  from nano to macro}, pages 720--723. IEEE, 2007.

\bibitem{chiu2013fast}
Liang-Chi Chiu, Tian-Sheuan Chang, Jiun-Yen Chen, and Nelson Yen-Chung Chang.
\newblock Fast sift design for real-time visual feature extraction.
\newblock {\em IEEE Transactions on Image Processing}, 22(8):3158--3167, 2013.

\bibitem{dalal2005histograms}
Navneet Dalal and Bill Triggs.
\newblock Histograms of oriented gradients for human detection.
\newblock In {\em 2005 IEEE computer society conference on computer vision and
  pattern recognition (CVPR'05)}, volume~1, pages 886--893. Ieee, 2005.

\bibitem{sift_mesh}
Tal Darom and Yosi Keller.
\newblock Scale-invariant features for 3-d mesh models.
\newblock {\em IEEE Transactions on Image Processing}, 21(5):2758--2769, 2012.

\bibitem{ppf}
Haowen Deng, Tolga Birdal, and Slobodan Ilic.
\newblock Ppf-foldnet: Unsupervised learning of rotation invariant 3d local
  descriptors.
\newblock In {\em Proceedings of the European conference on computer vision
  (ECCV)}, pages 602--618, 2018.

\bibitem{self_corner_detection}
Daniel DeTone, Tomasz Malisiewicz, and Andrew Rabinovich.
\newblock Superpoint: Self-supervised interest point detection and description.
\newblock In {\em Proceedings of the IEEE conference on computer vision and
  pattern recognition workshops}, pages 224--236, 2018.

\bibitem{maximum-weight-matching}
Ran Duan and Seth Pettie.
\newblock Linear-time approximation for maximum weight matching.
\newblock {\em Journal of the ACM (JACM)}, 61(1):1--23, 2014.

\bibitem{FSOD}
Qi Fan, Wei Zhuo, Chi-Keung Tang, and Yu-Wing Tai.
\newblock Few-shot object detection with attention-rpn and multi-relation
  detector.
\newblock In {\em CVPR}, 2020.

\bibitem{cnn-sift}
Philipp Fischer, Alexey Dosovitskiy, and Thomas Brox.
\newblock Descriptor matching with convolutional neural networks: a comparison
  to sift.
\newblock {\em arXiv preprint arXiv:1405.5769}, 2014.

\bibitem{nerf2nerf}
Lily Goli, Daniel Rebain, Sara Sabour, Animesh Garg, and Andrea Tagliasacchi.
\newblock nerf2nerf: Pairwise registration of neural radiance fields, 2022.

\bibitem{matchnet}
Xufeng Han, Thomas Leung, Yangqing Jia, Rahul Sukthankar, and Alexander~C Berg.
\newblock Matchnet: Unifying feature and metric learning for patch-based
  matching.
\newblock In {\em Proceedings of the IEEE conference on computer vision and
  pattern recognition}, pages 3279--3286, 2015.

\bibitem{local-ap}
Kun He, Yan Lu, and Stan Sclaroff.
\newblock Local descriptors optimized for average precision.
\newblock In {\em Proceedings of the IEEE conference on computer vision and
  pattern recognition}, pages 596--605, 2018.

\bibitem{horn}
Berthold K.~P. Horn.
\newblock Closed-form solution of absolute orientation using unit quaternions.
\newblock {\em J. Opt. Soc. Am. A}, 4(4):629--642, Apr 1987.

\bibitem{NERF-RPN}
Benran Hu, Junkai Huang, Yichen Liu, Yu-Wing Tai, and Chi-Keung Tang.
\newblock Nerf-rpn: A general framework for object detection in nerfs, 2022.

\bibitem{ke2004pca}
Yan Ke and Rahul Sukthankar.
\newblock Pca-sift: A more distinctive representation for local image
  descriptors.
\newblock In {\em Proceedings of the 2004 IEEE Computer Society Conference on
  Computer Vision and Pattern Recognition, 2004. CVPR 2004.}, volume~2, pages
  II--II. IEEE, 2004.

\bibitem{kim2022visual}
Hyunjin Kim, Minkyeong Song, Daekyeong Lee, and Pyojin Kim.
\newblock Visual-inertial odometry priors for bundle-adjusting neural radiance
  fields.
\newblock In {\em 2022 22nd International Conference on Control, Automation and
  Systems (ICCAS)}, pages 1131--1136. IEEE, 2022.

\bibitem{ExMeshCNN}
Seonggyeom Kim and Dong-Kyu Chae.
\newblock Exmeshcnn: An explainable convolutional neural network architecture
  for 3d shape analysis.
\newblock KDD '22, page 795–803, New York, NY, USA, 2022. Association for
  Computing Machinery.

\bibitem{adam}
Diederik~P. Kingma and Jimmy Ba.
\newblock Adam: A method for stochastic optimization, 2014.

\bibitem{kumar2009predicting}
Prateek Kumar, Steven Henikoff, and Pauline~C Ng.
\newblock Predicting the effects of coding non-synonymous variants on protein
  function using the sift algorithm.
\newblock {\em Nature protocols}, 4(7):1073--1081, 2009.

\bibitem{lin2021barf}
Chen-Hsuan Lin, Wei-Chiu Ma, Antonio Torralba, and Simon Lucey.
\newblock Barf: Bundle-adjusting neural radiance fields.
\newblock In {\em Proceedings of the IEEE/CVF International Conference on
  Computer Vision}, pages 5741--5751, 2021.

\bibitem{liu2008sift}
Ce Liu, Jenny Yuen, Antonio Torralba, Josef Sivic, and William~T Freeman.
\newblock Sift flow: Dense correspondence across different scenes.
\newblock In {\em Computer Vision--ECCV 2008: 10th European Conference on
  Computer Vision, Marseille, France, October 12-18, 2008, Proceedings, Part
  III 10}, pages 28--42. Springer, 2008.

\bibitem{hard-net}
Anastasiia Mishchuk, Dmytro Mishkin, Filip Radenovic, and Jiri Matas.
\newblock Working hard to know your neighbor's margins: Local descriptor
  learning loss.
\newblock {\em Advances in neural information processing systems}, 30, 2017.

\bibitem{morel2009asift}
Jean-Michel Morel and Guoshen Yu.
\newblock Asift: A new framework for fully affine invariant image comparison.
\newblock {\em SIAM journal on imaging sciences}, 2(2):438--469, 2009.

\bibitem{mueller2022instant}
Thomas M\"uller, Alex Evans, Christoph Schied, and Alexander Keller.
\newblock Instant neural graphics primitives with a multiresolution hash
  encoding.
\newblock {\em ACM Trans. Graph.}, 41(4):102:1--102:15, July 2022.

\bibitem{2dsift}
Pauline~C Ng and Steven Henikoff.
\newblock Sift: Predicting amino acid changes that affect protein function.
\newblock {\em Nucleic acids research}, 31(13):3812--3814, 2003.

\bibitem{park2017colored}
Jaesik Park, Qian-Yi Zhou, and Vladlen Koltun.
\newblock Colored point cloud registration revisited.
\newblock In {\em Proceedings of the IEEE international conference on computer
  vision}, pages 143--152, 2017.

\bibitem{peat2022zero}
Casey Peat, Oliver Batchelor, Richard Green, and James Atlas.
\newblock Zero nerf: Registration with zero overlap.
\newblock {\em arXiv preprint arXiv:2211.12544}, 2022.

\bibitem{anisotropic}
P. Perona and J. Malik.
\newblock Scale-space and edge detection using anisotropic diffusion.
\newblock {\em IEEE Transactions on Pattern Analysis and Machine Intelligence},
  12(7):629--639, 1990.

\bibitem{sift_point}
Yingying Ran and Xiaobin Xu.
\newblock Point cloud registration method based on sift and geometry feature.
\newblock {\em Optik}, 203:163902, 2020.

\bibitem{hypersim}
Mike Roberts, Jason Ramapuram, Anurag Ranjan, Atulit Kumar, Miguel~Angel
  Bautista, Nathan Paczan, Russ Webb, and Joshua~M. Susskind.
\newblock {Hypersim}: {A} photorealistic synthetic dataset for holistic indoor
  scene understanding.
\newblock In {\em International Conference on Computer Vision (ICCV) 2021},
  2021.

\bibitem{rublee2011orb}
Ethan Rublee, Vincent Rabaud, Kurt Konolige, and Gary Bradski.
\newblock Orb: An efficient alternative to sift or surf.
\newblock In {\em 2011 International conference on computer vision}, pages
  2564--2571. Ieee, 2011.

\bibitem{rusu2009fast}
Radu~Bogdan Rusu, Nico Blodow, and Michael Beetz.
\newblock Fast point feature histograms (fpfh) for 3d registration.
\newblock In {\em 2009 IEEE international conference on robotics and
  automation}, pages 3212--3217. IEEE, 2009.

\bibitem{mesh_dim}
Arman Savran and Bulent Sankur.
\newblock Non-rigid registration of 3d surfaces by deformable 2d triangular
  meshes.
\newblock In {\em 2008 IEEE Computer Society Conference on Computer Vision and
  Pattern Recognition Workshops}, pages 1--6, 2008.

\bibitem{des-compare}
Johannes~L Schonberger, Hans Hardmeier, Torsten Sattler, and Marc Pollefeys.
\newblock Comparative evaluation of hand-crafted and learned local features.
\newblock In {\em Proceedings of the IEEE conference on computer vision and
  pattern recognition}, pages 1482--1491, 2017.

\bibitem{3dsift}
Paul Scovanner, Saad Ali, and Mubarak Shah.
\newblock A 3-dimensional sift descriptor and its application to action
  recognition.
\newblock In {\em Proceedings of the 15th ACM international conference on
  Multimedia}, pages 357--360, 2007.

\bibitem{neural_descriptor_fields}
Anthony Simeonov, Yilun Du, Andrea Tagliasacchi, Joshua~B. Tenenbaum, Alberto
  Rodriguez, Pulkit Agrawal, and Vincent Sitzmann.
\newblock Neural descriptor fields: Se(3)-equivariant object representations
  for manipulation.
\newblock {\em CoRR}, abs/2112.05124, 2021.

\bibitem{Block-nerf}
Matthew Tancik, Vincent Casser, Xinchen Yan, Sabeek Pradhan, Ben Mildenhall,
  Pratul~P. Srinivasan, Jonathan~T. Barron, and Henrik Kretzschmar.
\newblock Block-nerf: Scalable large scene neural view synthesis, 2022.

\bibitem{Mega-nerf}
Haithem Turki, Deva Ramanan, and Mahadev Satyanarayanan.
\newblock Mega-nerf: Scalable construction of large-scale nerfs for virtual
  fly-throughs, 2021.

\bibitem{triplet}
Daniel~Ponsa Vassileios~Balntas, Edgar~Riba and Krystian Mikolajczyk.
\newblock Learning local feature descriptors with triplets and shallow
  convolutional neural networks.
\newblock In Edwin R.~Hancock Richard C.~Wilson and William A.~P. Smith,
  editors, {\em Proceedings of the British Machine Vision Conference (BMVC)},
  pages 119.1--119.11. BMVA Press, September 2016.

\bibitem{sift_rgb}
Jun Wan, Qiuqi Ruan, Wei Li, Gaoyun An, and Ruizhen Zhao.
\newblock 3d smosift: three-dimensional sparse motion scale invariant feature
  transform for activity recognition from rgb-d videos.
\newblock {\em Journal of Electronic Imaging}, 23(2):023017--023017, 2014.

\bibitem{margin_ranking_loss}
Jiang Wang, Yang song, Thomas Leung, Chuck Rosenberg, Jinbin Wang, James
  Philbin, Bo Chen, and Ying Wu.
\newblock Learning fine-grained image similarity with deep ranking, 2014.

\bibitem{kernel-sp}
Xing Wei, Yue Zhang, Yihong Gong, and Nanning Zheng.
\newblock Kernelized subspace pooling for deep local descriptors.
\newblock In {\em Proceedings of the IEEE conference on computer vision and
  pattern recognition}, pages 1867--1875, 2018.

\bibitem{wohlkinger2011ensemble}
Walter Wohlkinger and Markus Vincze.
\newblock Ensemble of shape functions for 3d object classification.
\newblock In {\em 2011 IEEE international conference on robotics and
  biomimetics}, pages 2987--2992. IEEE, 2011.

\bibitem{s_sift}
Jianchao Yang, Kai Yu, and Thomas Huang.
\newblock Supervised translation-invariant sparse coding.
\newblock In {\em 2010 IEEE Computer Society Conference on Computer Vision and
  Pattern Recognition}, pages 3517--3524. IEEE, 2010.

\bibitem{3dfeat}
Zi~Jian Yew and Gim~Hee Lee.
\newblock 3dfeat-net: Weakly supervised local 3d features for point cloud
  registration.
\newblock In {\em Proceedings of the European conference on computer vision
  (ECCV)}, pages 607--623, 2018.

\bibitem{3dmatch}
Andy Zeng, Shuran Song, Matthias Nie{\ss}ner, Matthew Fisher, Jianxiong Xiao,
  and Thomas Funkhouser.
\newblock 3dmatch: Learning local geometric descriptors from rgb-d
  reconstructions.
\newblock In {\em Proceedings of the IEEE conference on computer vision and
  pattern recognition}, pages 1802--1811, 2017.

\bibitem{zhang2017learning}
Xu Zhang, Felix~X Yu, Sanjiv Kumar, and Shih-Fu Chang.
\newblock Learning spread-out local feature descriptors.
\newblock In {\em Proceedings of the IEEE international conference on computer
  vision}, pages 4595--4603, 2017.

\end{thebibliography}
}

\end{document}